\documentclass[sigconf,10pt]{acmart}

\usepackage{enumitem}
\usepackage{subcaption}
\usepackage{balance}
\usepackage[noend,ruled]{algorithm2e}

\acmYear{2023}\copyrightyear{2023}
\setcopyright{rightsretained}
\acmConference[ACM MobiCom '23]{The 29th Annual International Conference on Mobile Computing and Networking}{October 2--6, 2023}{Madrid, Spain}
\acmBooktitle{The 29th Annual International Conference on Mobile Computing and Networking (ACM MobiCom '23), October 2--6, 2023, Madrid, Spain}
\acmPrice{}
\acmDOI{10.1145/3570361.3592523}
\acmISBN{978-1-4503-9990-6/23/10}

\begin{document}


\newcommand*{\splitl}[1]{%
\begingroup
    \renewcommand*{\arraystretch}{1.1}%
    \begin{tabular}[c]{@{}c@{}}#1\end{tabular}%
  \endgroup
}


\newcommand*{\splitr}[1]{%
\begingroup
    \renewcommand*{\arraystretch}{1.1}%
    \begin{tabular}[c]{@{}c@{}}#1\end{tabular}%
  \endgroup
}

\newcommand{\xref}[1]{\S\ref{#1}}

\newcommand{\squishlist}{\begin{itemize}[itemsep=1pt,parsep=2pt,topsep=3pt,partopsep=0pt,leftmargin=0em, itemindent=1em,labelwidth=1em,labelsep=0.5em]}
\newcommand{\squishend}{\end{itemize}}

\newcommand{\squishenum}{\begin{enumerate}[itemsep=1pt,parsep=2pt,topsep=3pt,partopsep=0pt,leftmargin=0em,listparindent=1.5em,labelwidth=1em,labelsep=0.5em]}
\newcommand{\squishsubenum}{\begin{enumerate}[itemsep=1pt,parsep=2pt,topsep=0pt,partopsep=0pt,leftmargin=0em,listparindent=1.5em,labelwidth=1em,labelsep=0.5em]}
\newcommand{\squishenumend}{\end{enumerate}}

\title{NeuriCam: Key-Frame Video Super-Resolution and Colorization for IoT Cameras}


\author{Bandhav Veluri, Collin Pernu, Ali Saffari, Joshua Smith, Michael Taylor, \\Shyamnath Gollakota}
\affiliation{%
  \institution{Paul G. Allen School of Computer Science \& Engineering, University of Washington}
  \city{\{bandhav, jrs, profmbt, gshyam\}@cs.washington.edu, \{cpernu, saffaria\}@uw.edu}
  \country{}
}

\renewcommand{\shortauthors}{Veluri, et al.}

\begin{abstract}
We present NeuriCam, a novel deep learning-based system to achieve video capture from low-power dual-mode IoT camera systems.  Our idea is to design a   dual-mode camera system where  the first mode  is low power (1.1~mW) but only outputs grey-scale, low resolution and noisy video and the second mode consumes much higher power (100~mW) but outputs color and higher resolution images. To reduce total energy consumption, we heavily duty cycle the high  power mode to output an image only once every second. The data for this  camera system is then wirelessly sent to a nearby plugged-in gateway, where we run our real-time  neural network decoder to reconstruct a higher-resolution color video.
To achieve this, we introduce an attention feature filter mechanism that
assigns different weights to different features, based on the correlation between the feature map and the contents of the input frame at each spatial location. 
We design a wireless hardware prototype using off-the-shelf cameras and address practical issues including packet loss and perspective mismatch. Our  evaluations show that our dual-camera approach reduces energy consumption by 7x compared to existing systems. Further, our model achieves an average greyscale PSNR gain of 3.7~dB over prior {single and dual-camera} video super-resolution methods and 5.6~dB RGB gain over prior  color propagation methods. 

 Open-source code: {\textcolor{blue}{{{\url{https://github.com/vb000/NeuriCam}}}}}. 
\end{abstract}



\maketitle

\section{Introduction}

\begin{figure}[!t]
\centering
    \includegraphics[width=\linewidth]{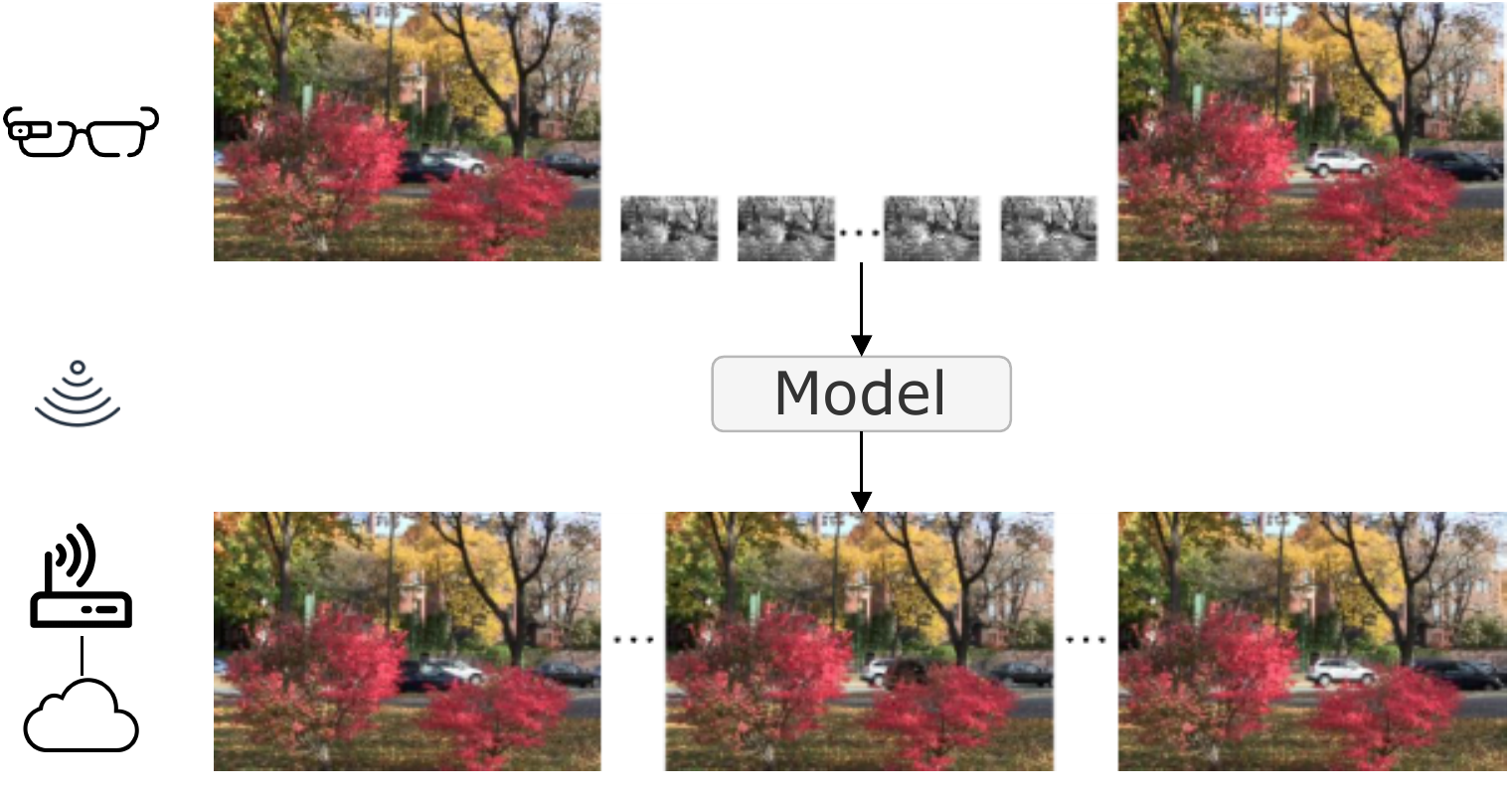}
    \caption{{\bf NeuriCam system.} Our dual-mode IoT camera system captures  low-resolution  gray-scale video from a low-power camera and  reconstructs high-resolution color frames using heavily duty-cycled high-resolution key-frames. The real-time neural network runs on the cloud or an edge server that is not power constrained.}
    \label{fig:fig1}
   \vskip -0.15in
\end{figure}

The power consumption of video camera systems is a bottleneck  for multiple applications like security cameras~\cite{sachin1, starfish, batteryvideo1,batteryvideo2}, wearable cameras,  robotics and sensor deployments  for wildlife  monitoring, and smart farms~\cite{wilds,chandra2021smart}. Recent works have demonstrated low-power  camera systems using off-the-shelf components but are limited to 1-5 fps and 160x120 greyscale  resolution~\cite{sachin1,sciencerobotics_uw}. 

While advances in sensor hardware have enabled microphones, temperature, and pressure sensors that are low-power (10--100 uW)~\cite{ekhonet,powifi},  camera  hardware can generate significantly more data and thus have   orders of magnitude higher power consumption~\cite{locamera-spec,hicamera-spec}. There is roughly a 100x difference in the power consumption of grey-scale, noisy, low-resolution image sensors and higher resolution RGB sensors (Fig.~\ref{fig:power_graph}) --- the HM01B0 sensor supports grey-scale, noisy QQVGA (160x120) resolution and  consumes  1.1~mW, while the OV7735 color sensor consumes 100~mW at VGA (640x480) resolution~\cite{locamera-spec,hicamera-spec}. This is due to three key factors. 
\squishlist
\item {\it Video resolution.} In contrast to  a temperature sensor that outputs a single value, a VGA camera  outputs 640×480 pixels per frame. At a  frame rate of 15 fps, this is around 4.6 million pixels per second. Thus, increasing the video resolution requires both a higher clock rate and significantly more energy consumption   to support the I/O operations~\cite{videoio,powerscaling}.
\item {\it Supporting color.} RGB image sensors that can support color require  increasing the number of channels per pixel which further increases the clock rate and I/O requirements.
\item {\it Noise and dynamic range.}  CMOS image sensors can reduce power consumption at the cost of higher read noise and  fixed pattern noise. This however results in a lower signal-to-noise ratio (SNR) and dynamic range~\cite{Liu_cmosimage,cmosnoise}. 
\squishend


\begin{figure}
    \centering
    \includegraphics[width=0.85\linewidth]{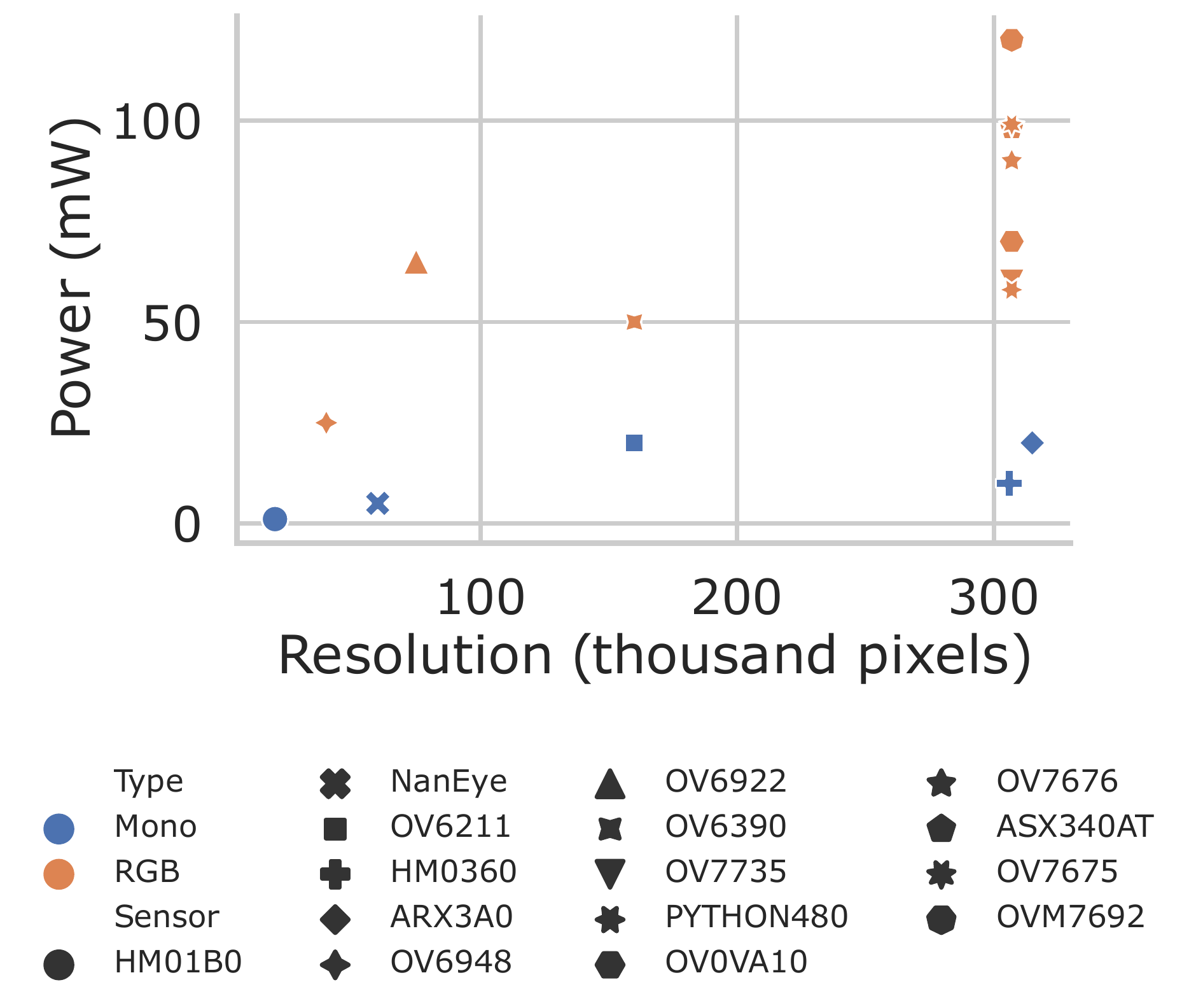}
    \vskip -0.15in
    \caption{
    {Power v/s resolution  of commercial  low-power image sensors.}
    While the list is not exhaustive, it shows the  relationship between power \& resolution.
    }
    \label{fig:power_graph}
    \vskip -0.2in
\end{figure}

This paper  explores the following problem: can we design high-resolution  color video camera systems while minimizing  camera energy  consumption? Our idea is to design a  dual-mode camera system where  the first mode is a low power (1.1~mW) but only outputs grey-scale, low resolution, and noisy video and the second mode consumes much higher power (100~mW) but outputs color and higher resolution images. To reduce total energy consumption, we heavily duty cycle the high-power camera to output an image only once every second. This data can  then be wirelessly sent to a nearby plugged-in access point or router that is not power constrained, where we run our real-time   neural network decoder to reconstruct a higher-resolution color video.

Our dual-mode approach to low-power video capture has three key advantages. First, heavily duty cycling the high-power image sensor reduces its energy consumption. For example, duty-cycling it to capture one frame per second, can reduce its average power consumption  by a factor of the target frame rate (e.g., a power reduction of  F times at F fps). Second, capturing lower-resolution monochrome video for the remaining frames can significantly reduce the amount of data captured by the camera system. For example, capturing greyscale 160x120 video at 15 fps and the key frame instead of RGB 640x480 video reduces the amount of data captured by a factor of 48. Finally, the router can combine  data across the two streams by super-resolving the low-resolution grey-scale video using the heavily duty-cycled  high-resolution video frames, which we refer to as key-frames. We show that these  key-frames  act as ground-truth priors for reconstructing precise high-frequency details that are missing in the low-resolution frames. In addition, these key-frames are also a  source of information for color, allowing us to generalize the super-resolution task to color.


Reconstructing high-resolution color video from our low-power dual-camera\footnote{Ideally, one could  design a custom camera IC that can   read out low-resolution grayscale continuously and provide a higher resolution color frame periodically with minimal switching delay. In practice, given the switching delays and resolutions of commodity low-power cameras, we implement the dual modes using two different cameras. In the rest of the paper, we reuse dual-mode and dual-camera interchangeably, given this context.} system is challenging for three reasons: (i) since the input video from the always-ON camera is grey-scale, key-frames from the heavily duty-cycled color camera are the only source of color. One na\"ive approach is to use the closest key-frame to propagate color across the video. However, occlusions in the current frame may not always be present in the closest key-frame due to motion, (ii) since we  capture the  higher resolution color key-frames only once per second, there can be significant relative motion between the objects in the key frame and current frame.  Standard convolutional  operations mostly focus on local features and if there is a significant relative motion between these frames, they tend to fail to capture features that may have moved hundreds of pixels away, and (iii)  since in our practical implementation, we use two cameras in slightly different locations, we can not assume that the transformation between the two views will preserve parallel lines. Thus, we need to account for different perspectives between the two cameras as well as   different fields of view while combining the data.
 

We present a novel deep learning-based system to achieve  video capture from low-power  IoT  systems. Our design uses a bidirectional recurrent network architecture that propagates information between the key-frames. The bidirectional nature of our network ensures that the two high-resolution key-frames that appear before and after the low-resolution frames are both used to super-resolve and colorize the video. We introduce a novel attention feature filter mechanism that assigns different weights to different features, based on the
correlation between the feature map and the contents of the input frame at {each spatial location}. This ensures that correct features are used to propagate color and high-resolution details  across all the frames  (see~\xref{section:attention}). To address the problem of high mobility across frames, we use optical flow-based alignment as an initial step to do long-range motion alignment and then refine it using deformable convolutions. We also aggregate temporal information to extract  details from neighboring frames  using grid propagation (see~\xref{sec:details}). Finally, we use a  fixed homographic transformation method to match the perspective of the  low-resolution camera to that of the high-resolution camera  (\xref{sec:practical}).


 We first evaluate our neural network using standard data sets to emulate a dual-mode system with different resolutions, which we train using the Vimeo-90K  dataset and perform the evaluation on the Vid4, UDM10 and REDS4 datasets. The REDS4 dataset is a first-person point-of-view camera dataset with a lot of random motion. Our evaluation shows:
\squishlist
\item Compared to state-of-the-art video super-resolution techniques (BasicVSR++~\cite{chan2021basicvsr}), our key-frame-based approach increases the gray-scale video PSNR by 1.7-4~dB across the three datasets (for fairness, our PSNR computations exclude the key-frames). This is achieved while using a comparable parameter size (7.3~MB) as BasicVSR++ (7~MB).

\begin{figure*}[t!]
    \centering
    \begin{subfigure}[b]{0.7\linewidth}
        \centering
        \includegraphics[width=\linewidth]{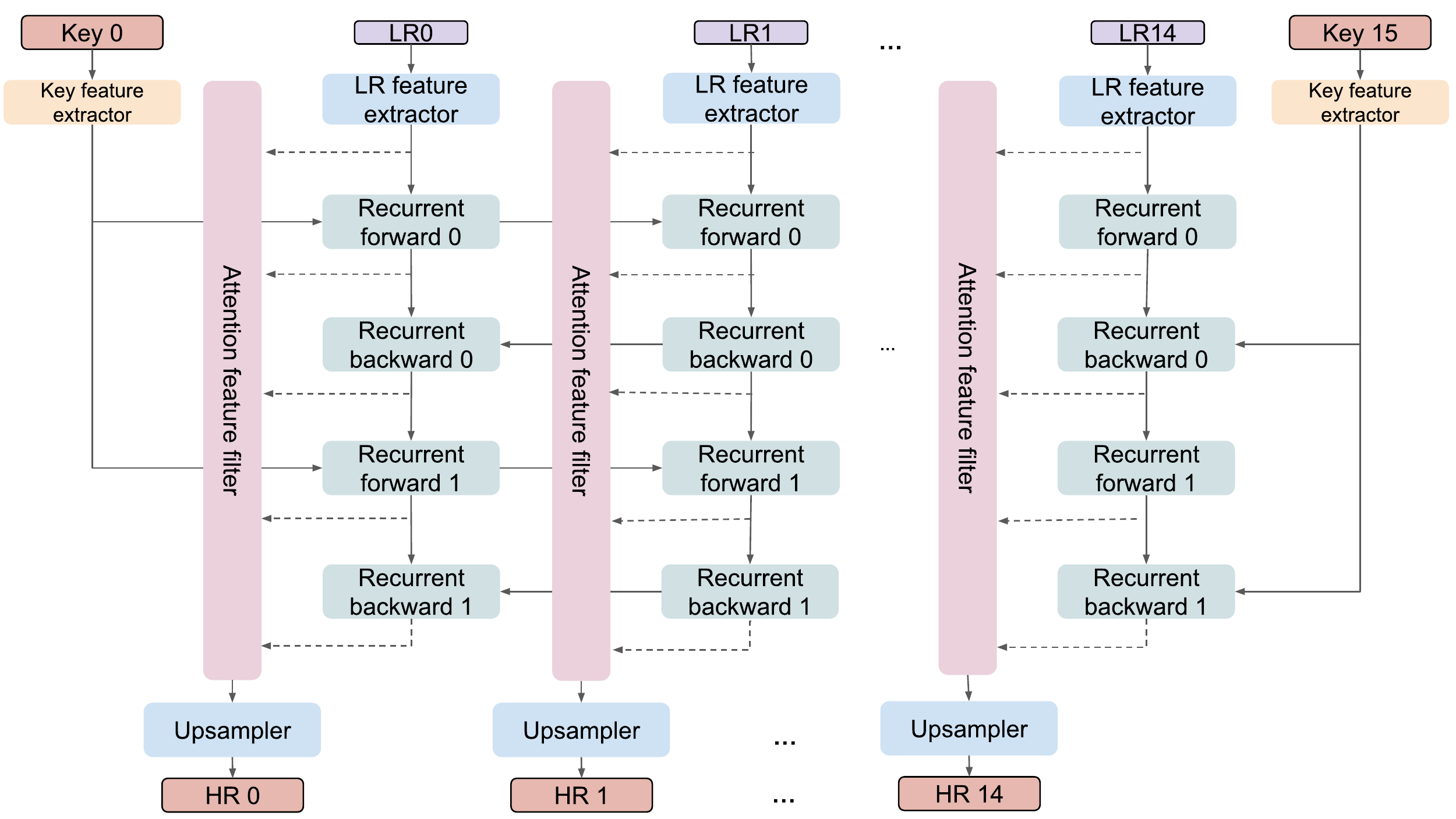}
    \end{subfigure}
    \hspace{2em}
    \begin{subfigure}[b]{0.165\linewidth}
        \centering
        \includegraphics[width=\linewidth]{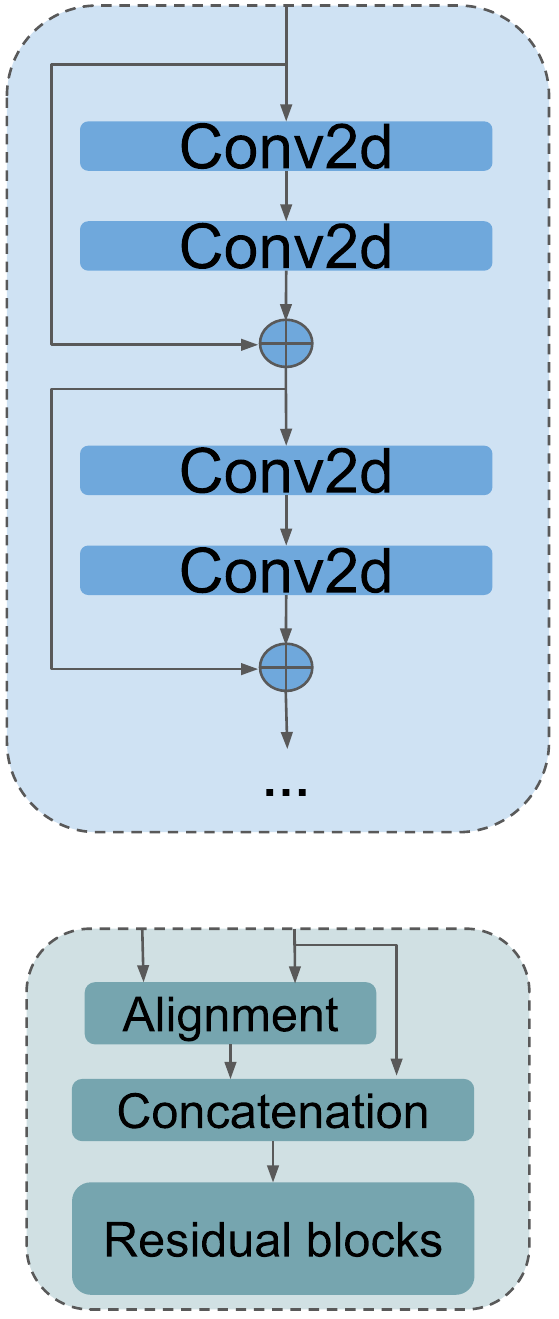}
    \end{subfigure}
    \caption{{\bf Our model  with aligned grid propagation and attention similarity filters.}
    Residual blocks (shown in blue in the right figure) are the processing elements used in feature extraction
    and feature propagation. Recurrent elements (green blocks) align the neighboring features
    and compute aggregated features at each level in the grid.}
    \label{fig:model}
    \vskip -0.1in
\end{figure*}

\item Compared to AWnet ~\cite{Cheng2021ADC}, our attention-based approach increases the grey-scale video PSNR by 3.6-7 dB. These improvements in the video quality can help with the more faithful reconstruction of finer details such as signs, letters, shapes, and textures (Fig.~\ref{fig:vsr_qual_comp}) and are possible because we can learn from the finer details in the high-resolution key-frames that are absent in the low-resolution video.\footnote{AWnet \cite{Cheng2021ADC} proposes a dual-camera system to achieve high frame rate video acquisition using a low-resolution high frame rate camera and a high-resolution low frame rate camera. This work differs from our work in the overall objective and the proposed solution. AWnet maximizes the output frame rate, while our work minimizes the overall power and bandwidth consumption. Furthermore, AWnet does not consider the colorization task.}
\item We achieve a 4-6~dB gain in  colorization performance compared to color propagation methods (DEVC~\cite{10.1145/3197517.3201365}). In comparison to reference-based image super-resolution methods (TTSR~\cite{inproceedings}), our  method  achieves a 4-6~dB and 4-7~dB improvement in Y and RGB channels, respectively.
\squishend

\noindent{\bf Hardware evaluation.} We prototype low-power dual-camera hardware using off-the-shelf components. Our prototype includes a Himax HM01B0 image sensor capturing greyscale video at 15~fps and QQVGA resolution  and  an OmniVision OV7692 color VGA sensor capturing key-frames once every second.  Our prototype design includes microcontrollers and wireless radios to support data transmission from these devices. Our hardware evaluation  shows that,
\squishlist
\item On the data collected from our dual-camera hardware, we achieve an average gain of 3.7~dB in the PSNR of greyscale video in comparison with BasicVSR++. Compared to re-trained TTSR, we see an RGB performance gain of 5.6~dB.
\item The runtime of our model on an Nvidia RTX
2080 Ti GPU is around 54 ms per frame. For context, at 15 fps, the receiver must process each frame in less than 66~ms to be real-time. By increasing the batch size and GPU utilization, we can reduce the per-frame processing time to 44~ms. 
\item Our  dual-camera hardware prototype achieves 640x480 color wireless video  at 15~fps while consuming only 46~mA. 
\item {We perform a design space analysis with on-device compression which shows that our approach achieves better/ comparable PSNRs than single-camera codec system at IoT bit rates while reducing the estimated sensor and codec power consumption by 8x and 7-8x respectively (\xref{sec:exploration}).}
\squishend

\vskip 0.05in\noindent{\bf Contributions.} We make the following contributions:
\squishlist
\item The first dual-mode camera  system design where the first mode is low power but outputs grey-scale, low resolution video while the second mode is high power but outputs heavily duty-cycled color and higher resolution images.
\item Motivated by our low-power system requirements,  we design an efficient deep learning technique for key-frame based video super resolution and colorization using an attention feature filter mechanism that propagates color and high resolution features across video frames.
\item Perform evaluations that show significant gains over prior video super resolution and  color propagation methods.  
\item Built prototype hardware addressing practical issues like  perspective mismatch and wireless packet losses  and show   energy savings while achieving reasonable resolution. \item  By making our code and data open source, we hope to help with reproducibility in  ML-based   video IoT systems. 
\squishend

\section{System Design}

\subsection{Design of the neural network model}

Fig.~\ref{fig:model} shows the architecture of our neural  model. We split the input
video sequence into key-frame to key-frame sequences before they are fed into the network. So, in each iteration,
the network operates on 15 low-resolution frames, $\{l_{t}\}_{t=0}^{t=14}$, and two key-frames (previous and next),
$\{k_{t}\}_{t=0}^{t=1}$.  While the bidirectional recurrent network with forward and backward paths as well as the feature extractors are building blocks we use from prior work, the attention feature filters are novel blocks we introduce in this
work for this task. 

\subsubsection{Attention feature filters}
\label{section:attention}
Since we extract and map high-resolution details and color from key-frames,
it is vulnerable to scenarios where an object appears in a frame between the key-frames, but the closest key-frame does not have the object. Since we get color/HR information primarily from the key-frames, consider a scenario where frame 2 has an object that appears in view, but is not present in the previous (closest) key frame 0 but is present in the next key frame. The recurrent forward blocks in Fig.~\ref{fig:model} are heavily influenced by the previous key frame since it is the closest key frame. The color/HR information for such an object might only be present in the next (farther) key frame. While alignment and propagation are effective at aggregating features from the entire sequence, those blocks do not explicitly consider the content of the frame and its correlation to features at each level. 
This results in blurred and grayed-out details,  limiting our ability to effectively super-resolve and colorize.

\vskip 0.05in\noindent{\bf Our attention mechanism.} To solve this problem, we propose a feature filter based on attention
mechanism \cite{Bahdanau2015NeuralMT}, that assign varying levels of weightage to each filter level, based on the
correlation between the feature map and contents of the input frame at \emph{each spatial location}.

For time steps $t \in \{0, 14\}$, let $w \times h$ be the resolution of the low-resolution input frames,
$f_t \in \mathbb{R}^{w \times h \times 64}$ be the low-resolution input feature map computed by the low-resolution feature
extractor, and  $f_{l,t} \in \mathbb{R}^{w \times h \times 64}$ be aggregated features output by alignment
and propagation blocks at level $l$ in the grid shown in Fig.~\ref{fig:model}. 64 is the channel depth of the feature maps that our network operates on. Then, the attention feature filter,
$\hat{f}_{t} \in \mathbb{R}^{w \times h \times 64}$, is given by:
\begin{equation} \label{eq1}
\hat{f}_{t} = \sum_{l=0}^3 f_{l,t} \odot A(f_{l,t}, f_t)
\end{equation}
where $ A(f_{l,t}, f_t) \in \mathbb{R}^{w \times h} $ is the attention score value (which effectively is the weight) between low-resolution input feature map
and the aggregated feature map at level $t$ at \emph{each spatial location}. And, $\odot$ effectively denotes the element-wise multiplication
between the feature map and attention value computed for that feature map at that spatial location where all the 64 values/channels in the feature map are multiplied by the attention value. Thus, the attention score value is unique for each spatial location in the frame but is the same across all 64 channels in the feature map. This allows us to uniquely sample the relevant feature maps at each spatial location, based on the contents of that location.

We use the following definition for the attention score function at each spatial location $(i, j)$:
\begin{equation} \label{eq2}
A(f_{l,t}, f_t) ^ {i,j} = Softmax(f_{l,t}^{i,j} \cdot f_t^{i,j})
\end{equation} 
Here, we first compute the dot product similarity between the low-resolution features, $f_t \in \mathbb{R}^{64}$, and aggregated features
$f_{l,t} \in \mathbb{R}^{64}$, at spatial location ${i, j}$. This results in four similarity scores, one for each propagation level. We use
these scores to choose the features map with maximum similarity. But since choosing the maximum would be a non-differentiable operation,
we use  $Softmax$ to perform non-maximum suppression of the scores in a differentiable way.


The importance of our attention mechanism to our task of key-frame-based super-resolution and colorization is
demonstrated in Fig.~\ref{fig:attn_comp}. Here, we can observe that the occluded region is better predicted by the model
with attention using information from the next key-frame, despite  frame 7 being closer in time, to the
previous key-frame. 

\begin{figure}
    \centering
    \begin{subfigure}[b]{\linewidth}
        \centering
        \includegraphics[width=\linewidth]{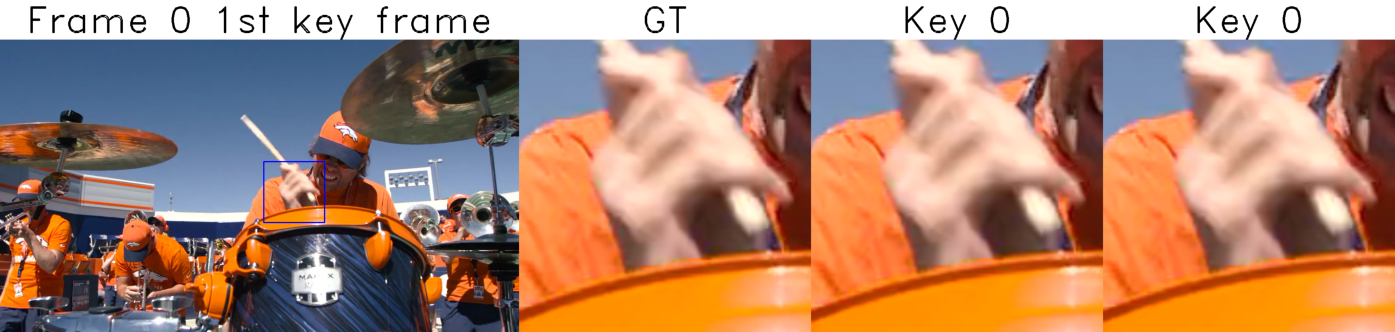}
    \end{subfigure}
    \begin{subfigure}[b]{\linewidth}
        \centering
        \includegraphics[width=\linewidth]{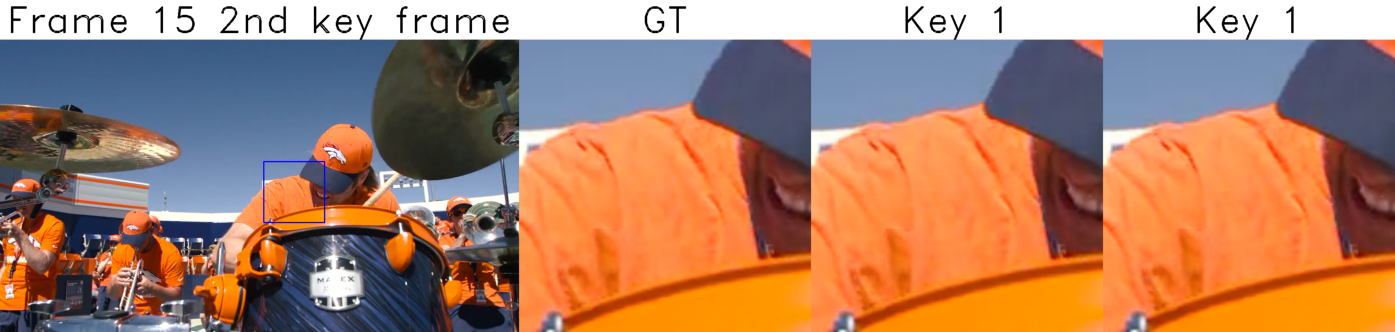}
    \end{subfigure}
    \begin{subfigure}[b]{\linewidth}
        \centering
        \includegraphics[width=\linewidth]{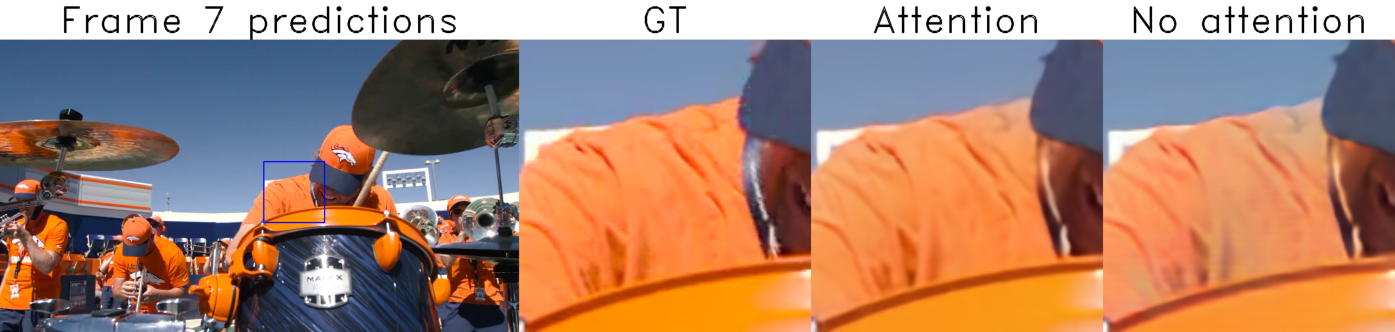}
    \end{subfigure}
    \vskip -0.1in
    \caption{{\bf Our attention mechanism on a fast motion video.} Top and
    middle pictures show the first and second key-frames. The bottom picture shows the prediction
    of frame 7 by our models with and without attention. The model with attention is able to infer details
    from the second key-frame due to its higher correlation, while the model without attention generates
    ambiguous color.}
    \vskip -0.15in
    \label{fig:attn_comp}
\end{figure}

\subsubsection{Neural network details for remaining blocks in Fig.~\ref{fig:model}}\label{sec:details}


\vskip 0.05in\noindent{\bf Feature extractors.} Both the low-resolution and high-resolution input frames are fed into feature extraction blocks. We use 5 residual blocks for the low-resolution feature extractor and 7 residual blocks for
key-frame feature extractor. As shown in Fig.~\ref{fig:model}, each residual block is a sequence of two 2D convolution operations followed by a residual skip connection. The first two convolutional layers in the key-frame feature extractor subsample the input spatially while increasing the channel depth to generate key-frame features that have the same dimensions as low-resolution features. We use separate feature extraction blocks with independent parameter sets for low-resolution
frames and key-frames. The feature extractors used for individual low-resolution frames, or individual key-frames,
are the same.

\vskip 0.05in\noindent{\bf Grid Propagation.} 
Aggregating temporal information is important for extracting details from neighboring frames and key-frames. We use grid propagation proposed in \cite{chan2022basicvsrplusplus} to achieve feature aggregation over the entire input sequence refined over multiple levels. As shown in the green blocks in Fig.~\ref{fig:model}, each cell in the grid performs two functions: \emph{alignment and aggregation}. In videos with high relative motion between subsequent frames, hidden states computed based on previous  frames might align poorly with the current frame. Since  convolution operations tend to fail at effectively considering  global correlations \cite{dosovitskiy2020vit, DBLP:journals/corr/abs-2110-02178}, it is hard for residual blocks to transform previous frame features to align with the current frame. \cite{tian2020tdan, wang2019edvr} proposes the use of deformable convolutions to mitigate this problem. Deformable convolutions are generally hard to train due to the instability caused by large filter offsets. To mitigate this, the alignment module uses optical flow-based alignment as an initial step to do long-range motion alignment and then refines it using deformable convolutions, which would now have reasonable offsets. Note that alignment and propagation modules at  each level are different (with unique parameter sets), but the same module is used across all time steps. The aligned features are then processed using a residual block (similar in architecture to that of feature extractors) to generate features with temporal context. For super-resolution tasks, this technique is shown to improve the performance by 0.5 dB \cite{chan2022basicvsrplusplus}. We use two levels of forward followed by backward processing resulting in a grid propagation of features.

\vskip 0.05in\noindent{\bf Upsampler.} Once we compute the aligned feature propagated from both directions, we apply the attention filter described in \xref{section:attention}
to obtain the final feature map, that we can predict the high-resolution color frames with. We use two convolutional layers to generate 48 
channel output, which is then followed by a pixel shuffle layer  \cite{7780576}  to upsample the output to a 3-channel high-resolution output. We apply $tanh$ activation to the final layer to convert the output to be in the $[-1, 1]$ range. The output we obtain is in L*a*b* colorspace,
so we normalize the output to $[0, 1]$ range, and convert it to RGB colorspace. We normalize the resulting  output to $[0, 255]$, round 
the values, and store them as  8-bit unsigned integers.

        

\vskip 0.05in\noindent{\bf Color space and loss function.}
Most prior super-resolution works operate in the RGB colorspace. However, since we are also performing colorization
tasks in tandem, using RGB colorspace resulted in subpar colorization results in our experiments. So, we convert the
key-frame, and during training ground-truths, to CIELAB colorspace, which has been shown to perform better for colorization
\cite{zhang2017real, 10.1145/3197517.3201365}. For the loss function, however, we found that Charbonnier loss \cite{413553}
popular in video super-resolution works better compared to L1 or Huber loss functions used in colorization works. Given
the  high-resolution output frame $HR_{predicted}$, and ground-truth high-resolution frame $HR_{gt}$, the Charbonnier
loss is given by, $\mathcal{L} = \sqrt{||HR_{predicted} -  HR_{gt}||^2 + \epsilon^2} $, where $\epsilon$ is the regularization factor which we set to $1e-3$ for training.

\subsection{Practical Issues}\label{sec:practical}




\noindent{\bf Perspective correction.} Since the two camera locations with respect to each other are fixed,  any transformation between  them would be a 
fixed transformation that can also be applied as a preprocessing step to the model's input. Assuming planar views of both cameras, we  estimate the homographic transformation of the two views.
Given four reference points in a low-resolution frame and key-frame, we  transform an arbitrary point $(x, y)$
in low-resolution plane to key-frame plane $(x_k, x_k)$, by:
\begin{equation}
    \begin{bmatrix}
    x_k \\
    y_k \\
    1
    \end{bmatrix}
    =
    \begin{bmatrix}
    h_{11} & h_{12} & h_{13} \\
    h_{21} & h_{22} & h_{23} \\
    h_{31} & h_{32} & 1
    \end{bmatrix}
    \begin{bmatrix}
    x \\
    y \\
    1
    \end{bmatrix}
\end{equation}


We estimate this fixed homographic matrix  using four known  reference points during a one-time calibration step. We then identify
the reference points in the low-resolution frame and key-frame frame. Using these points in low-resolution frame as the start, we estimate the above parameters.

\vskip 0.05in\noindent{\bf Packet losses.} In practice, video data transmitted wirelessly can be lost due to packet losses on the wireless channel. Retransmissions can increase the overhead on the network and increase power consumption.  In our experiments, we observe that packet loss translates to a loss of 1-2 image lines, on average. However in
adverse cases, multiple packets might be lost at a time, and our packet loss handling mechanism should handle that gracefully. We implement a packet correction mechanism, that first detects the start index and count of the number of lines lost
and stores it in a key-value map. Starting from the top index in the map, we consider extracting a 5-pixel tall snippet of the error-free image, 
immediately on top of the error line, and using bicubic interpolation to expand the snippet to cover the lost lines. This ensures that starting from the top we progressively fill the lost lines all the way to the bottom index. 

\subsection{On-device compression}\label{sec:exploration}

{Commercial IoT cameras use a single high-resolution sensor and employ on-device codec to reduce  the data rate to reasonable levels. In this section, we explore the application of our dual-mode sensing approach in such scenarios by evaluating our model with input streams compressed using h.264 codecs. In our setting, both the low-resolution grayscale video at 15 fps and the high-resolution color video at 1 fps were separately encoded using h.264. We use three-quarters of the bitrate for the compressed  low-resolution stream and use the rest for the compressed  high-resolution stream at 1 fps. Empirically, we found this ratio to yield the best results. We compare the quality of the output of our method with that of a single-camera codec-based system, that compresses the raw high-resolution video at 15 fps, using the same bitrate as the combined bitrate of the low-resolution and high-resolution 1 fps stream used for our method.  That way, we ensure the total bitrate used by our dual-mode codec approach and prior single-camera codec approach are  equal. Fig. \ref{fig:codec_comp_psnr} shows that our model shows better/comparable PSNR when fed with compressed input streams, in comparison with the performance of raw high-resolution video compressed at an equal bitrate. We observe that our method shows better performance at lower bitrates making it well-suited for IoT and wearable applications. This design characteristic also enables the use of ultra-low-power, but bit rate-limited communication approaches such as backscatter  where increasing the bit rate requirements significantly reduces the range \cite{10.1145/3130970}.}

{Given the same number of bits needed for communication (after compression), Fig.~\ref{fig:codec_comp_power} compares the power consumption of the sensor and compression in the two systems. For sensor power, we use the power consumption of low-resolution and high-resolution sensors described in \xref{sec:hardware}. Our dual-model sensing approach uses the lower-powered low-resolution grayscale sensor most of the time and uses the higher-powered high-resolution sensor only once every second for 40 ms. On the other hand, the single-camera approach uses a higher-powered high-resolution sensor at 15 fps which consumes 46.2~mW. As a result, our dual-mode approach consumes 8x lower sensor power compared to the single-camera approach. We estimate the power consumption of compression using two implementations of h.264 codec: an academic ultra-low-power hardware IP \cite{4914842} (HW IP), and a commercial off-the-shelf IC \cite{h264_ic} (COTS IC) implementation of h.264 codec. Since the 4x low-resolution stream contains 16x fewer pixels compared to the raw high-resolution stream, and the key-frame stream is heavily duty-cycled, the codec has much fewer  pixels when compared to compressing the raw high-resolution video. Since the  power consumed by a well-designed codec scales linearly with pixel count \cite{10.1145/3115505}, compression consumes 7-8x lower power with our approach. Note that in the above comparison, both systems use the same number of communication bits and hence have the same communication power consumption (which can be sub-mW and not an energy bottleneck  with  backscatter~\cite{dinesh}).}

\begin{figure}
    \centering
    \begin{subfigure}[b]{\linewidth}
        \centering
        \includegraphics[width=0.95\linewidth]{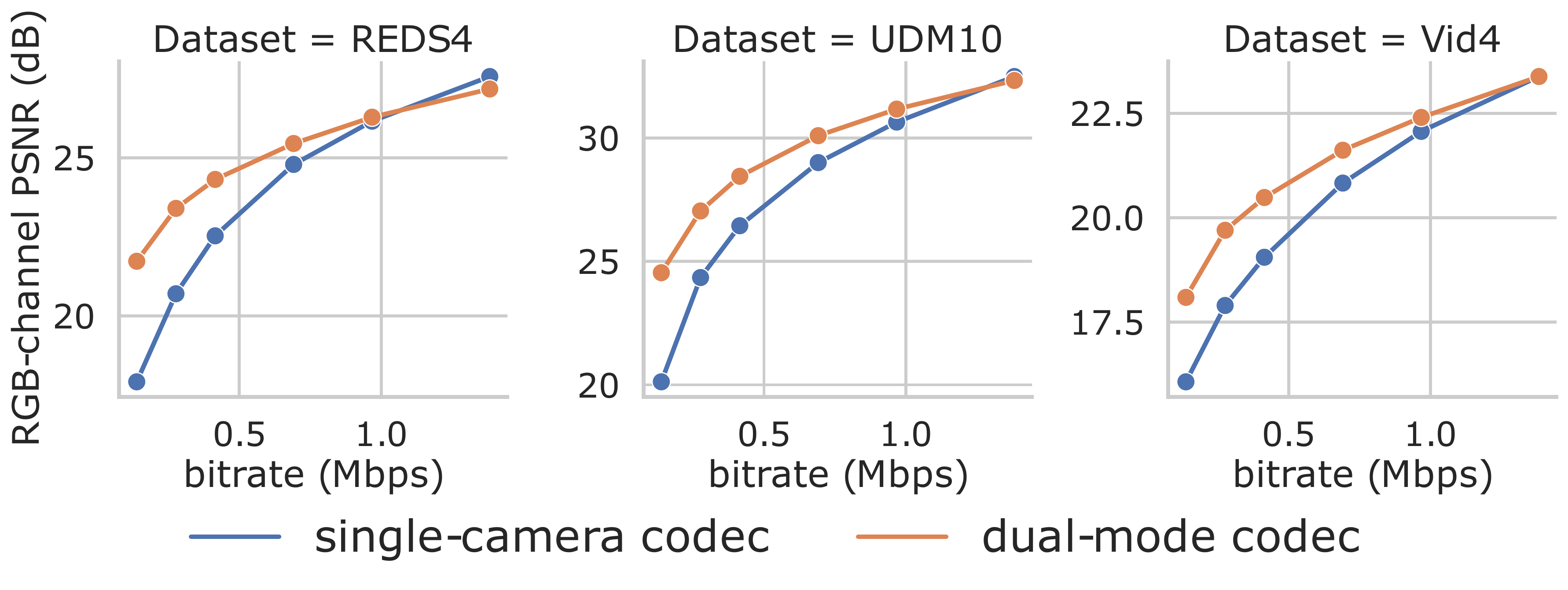}
        \vskip -0.1in
        \caption{Performance evaluation of our method with h.264 codec.}
        \label{fig:codec_comp_psnr}
    \end{subfigure}
    \begin{subfigure}[b]{\linewidth}
        \centering
        \includegraphics[width=0.95\linewidth]{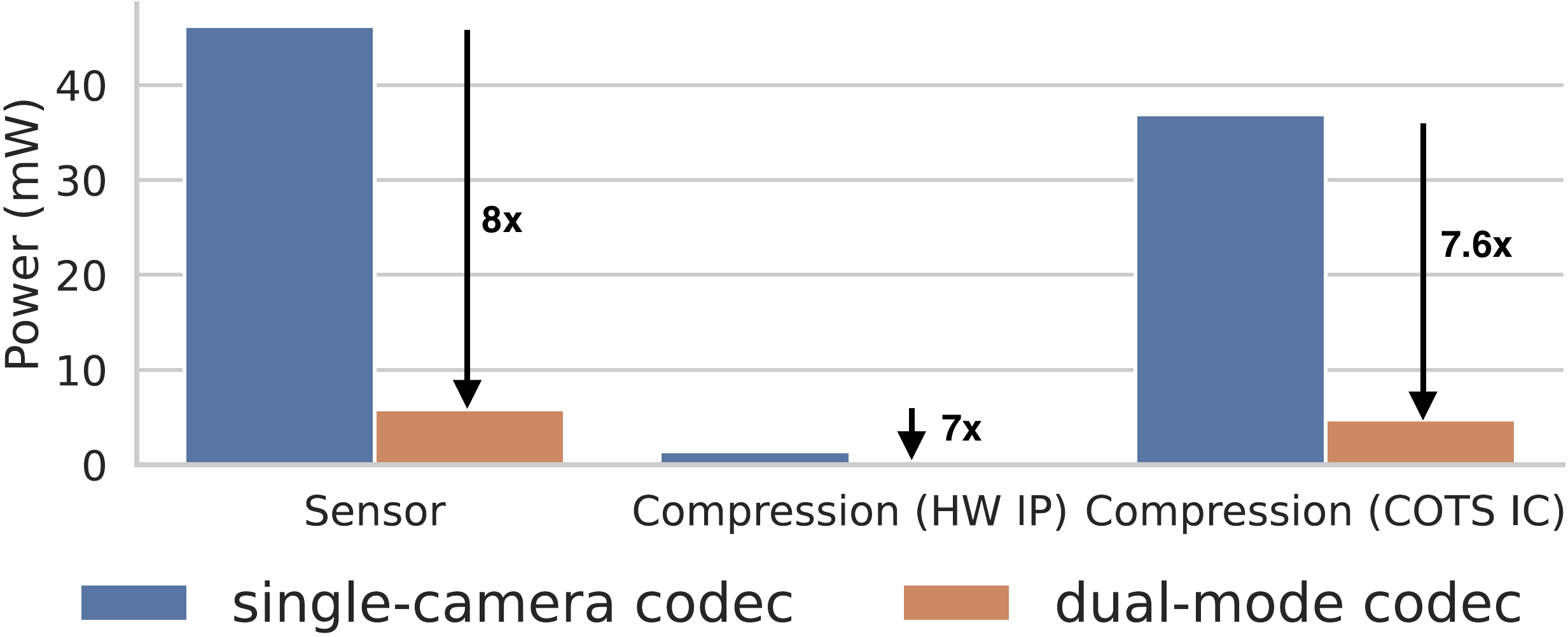}
        \caption{Comparison of sensor and compression power.}
        \label{fig:codec_comp_power}
    \end{subfigure}
    \vskip -0.1in
    \caption{When using on-device video codec, our dual-mode sensing approach results in better or comparable PSNR to traditional single-camera systems.}
    \vskip -0.15in
    \label{fig:codec_comp}
\end{figure}

\vskip 0.15in \subsection{Low-power dual-camera hardware}\label{sec:hardware}

\begin{figure}
    \centering
    \includegraphics[width=\linewidth]{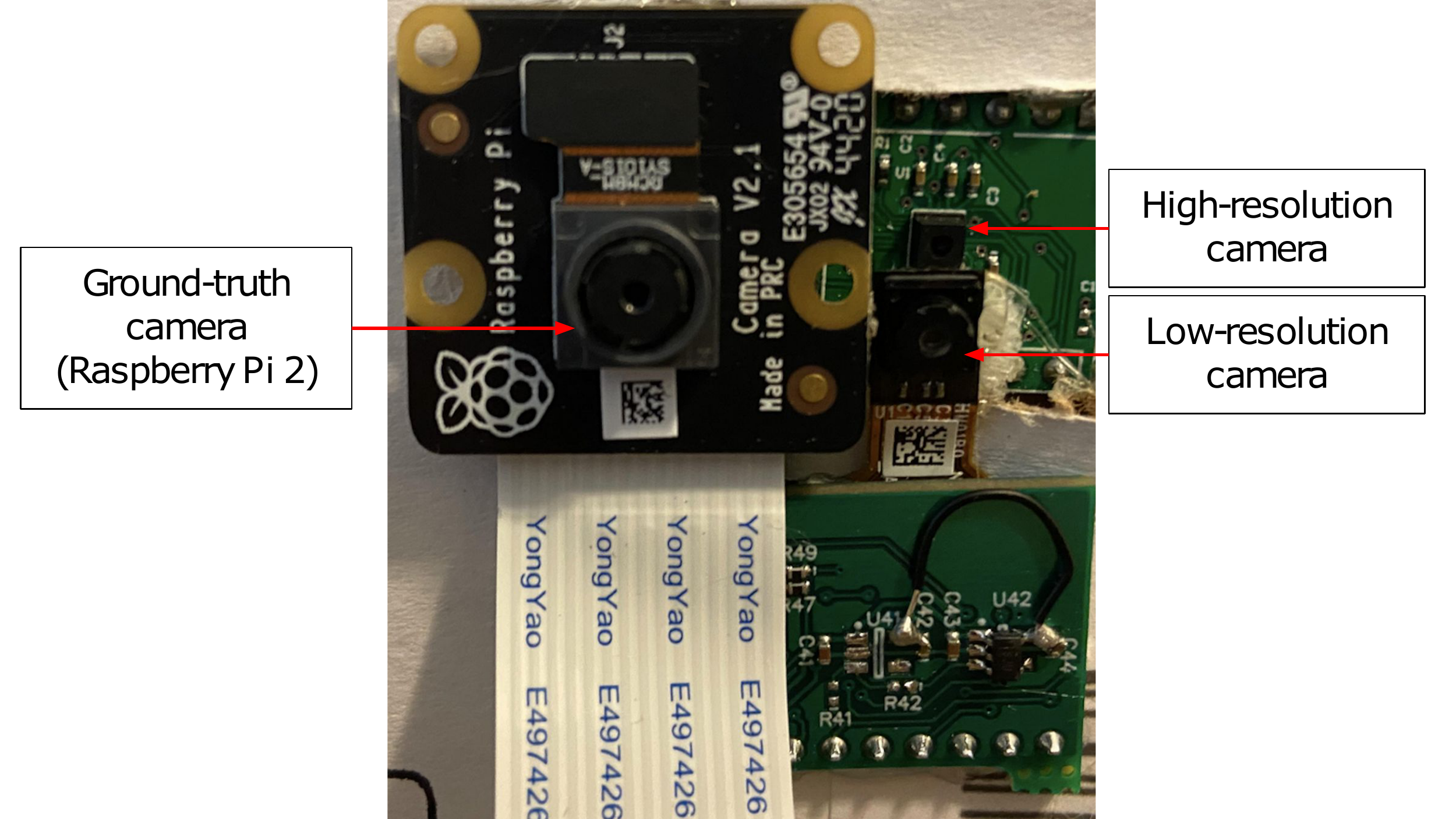}
    \vskip -0.1in
    \caption{\textbf{Prototype of the dual camera system.} We use Raspberry Pi cameras for  ground truth. The flex PCB cable allows us to super-glue the two cameras close by.}
    \label{fig:cameras}
    \vskip -0.2in
\end{figure}

{\bf Low-resolution grayscale camera system.} We use a low-power STM32L496 microcontroller (MCU) to initialize the Himax HM01B0 image sensor in
QQVGA mode and record 15 frames of grayscale video in one second. Our MCU operating at 80 MHz,
uses its PLL (Phase Locked Loop) module to provide a 10 MHz clock for the image sensor operation.
Once the image sensor clock is ready, the MCU uses the I2C module to initialize the image sensor in
video mode and send commands to the Himax camera.
Next, the MCU uses the SPI module in slave mode to receive the 8-bit pixels value from the image sensor
(SPI master). A DMA (Direct Memory Access) channel is enabled to transfer the data from the sensor
(peripheral) to the MCU (memory). This allows us to leave the MCU central core out of the reading
process and keep it in sleep mode, which results in lower power consumption.
Finally, to transfer the collected data to the basestation, we can use a TI CC2640R2F radio in 8-FSK mode
with a maximum throughput of 2.5 Mbps. The radios can be set to different frequencies across  at 2.4 MHz band and transmit at 0 dBm. Our STM32L496 MCU uses a DMA channel to transfer data to the
TI radio’s memory via UART. 

\vskip 0.05in\noindent{\bf High-resolution color camera system.} 
Our design uses a STM32U575 MCU plus an OmniVision OV7692 color VGA image sensor to record one key-
frame per second. The MCU, operating at 160 MHz, generates a 12 MHz clock signal for the image
sensor’s operation and initializes it in YUV 4:2:2 mode through an I2C connection. In this mode, the
image sensor outputs 16 bits of data per pixel, resulting in a frame data size of 5 Mb, which is smaller than our MCU’s RAM capacity. After the initialization process, the sensor captures a picture and uses its digital video port (DVP) parallel
output interface with 8 data lines and three synchronization signals (horizontal reference, vertical reference, and pixel clock) to transfer the image. A DMA channel of the microcontroller is activated to transmit 8 bits of data on the positive edge of the pixel clock signal until a frame is complete. The central MCU core is mainly kept in sleep mode during image reading to preserve energy. To reduce the energy consumption even more, the VGA image sensor is only active for 40ms which is the duration of the initialization and reading processes. Since the radio’s maximum throughput is 2.5 Mbps, we divide the data into two streams and use two separate TI radios operating at two different center frequencies to transfer the data. Our STM32U575 MCU can enable two DMA channels and two UART modules (one for each radio) to send the video data to the radios’ memory.

To synchronize the frames across the two cameras,  we add a specific three-byte footer at the end of every frame (13, 0, and 10) and transfer it alongside the image data. The base station looks for the first footer in the coming data to find the start pixel and reconstruct the images. We implement a 32-bit timer on the low and high-resolution MCUs with a precision of 1 ms. After capturing a frame, the MCU reads the 32-bit timer value and appends it to the image data.

For the training process to be most effective, we need the ground-truth camera used in  data collection to capture exactly the same scene as the low-resolution camera. This means that the ground-truth frames are both spatially and temporally aligned with the low-resolution frames. Spatial alignment is achieved by perspective correction as described in \xref{sec:practical}. Temporal alignment requires time synchronization of the cameras. In our system, each sensor is controlled by a separate microcontroller. We achieve synchronization using an interrupt generated by  the MCUs, which resets the counters running on all the microcontrollers at the same instant, achieving synchronization.

\section{Evaluation}


\subsection{Benchmarking our neural network}


\vskip 0.05in\noindent{\bf Training setup.} We use the Vimeo-90K \cite{xue2019video} dataset as the standard training dataset for
all the method we compare with. The Vimeo-90K dataset is a large collection of 39K videos with 7 frames per video. Given the limited number of frames per video, we use a key frame interval size of 6 for training (but 15 for evaluation). The sequences are selected such that each sequence
has sufficient motion. Following  \cite{10.1007/978-3-030-58610-2_38}, we pad the input frames  with 2 pixels, and
equivalently pad the key-frames  with 8 pixels, along the borders in reflect mode. This would result in an output
padded with 8 pixels along the borders, with any edge artifacts induced by the model limited to those 8 pixels.
Later, the padding on the output is removed as a post-processing step before evaluation. During training, we augment the
dataset by sampling a random crop of size 256x256 from a video sample and then by applying random flipping and
rotation transformations. Each video sample is converted to the L*a*b* color space and normalized to $[-1, 1]$ range.
We use Matlab's \texttt{imresize} in bicubic downsampling mode to obtain low-resolution input stream for evaluations
on standard datasets. We use Adam optimizer~\cite{article}  with $\beta_1=0.9$, $\beta_2=0.999$ and
$\epsilon=10^{-8}$. We train the model with a batch size of 8 for a total of 80 epochs, with
an initial learning rate set to $10^{-4}$. The learning rate is scheduled using \texttt{ReduceLROnPlateau}
learning rate scheduler with reduction factor and patience set  to 0.1 and 5 epochs.

\vskip 0.05in\noindent{\bf Evaluation setup.} We use three widely used data sets  for evaluation:
Vid4 \cite{6549107}, UDM10 \cite{PFNL} and REDS4 \cite{wang2019edvr}. Vid4 is a set of 4 videos with
varying resolutions and frame count, that contain fine textures that are hard for models to predict.
UDM10 is a set of 10 1280x720 videos, each containing 32 frames, that captures relatively high motion.
REDS4 is a first-person point-of-view dataset, and thus has a lot of random motion spanning the entire
video. REDS4 is set of 4 1280x720 videos each containing 100 frames. The key frame interval is set to 15
matching that of the hardware system. 

\begin{figure}[t]
    \centering
    \begin{subfigure}[b]{0.4\linewidth}
        \centering
        \includegraphics[width=\linewidth]{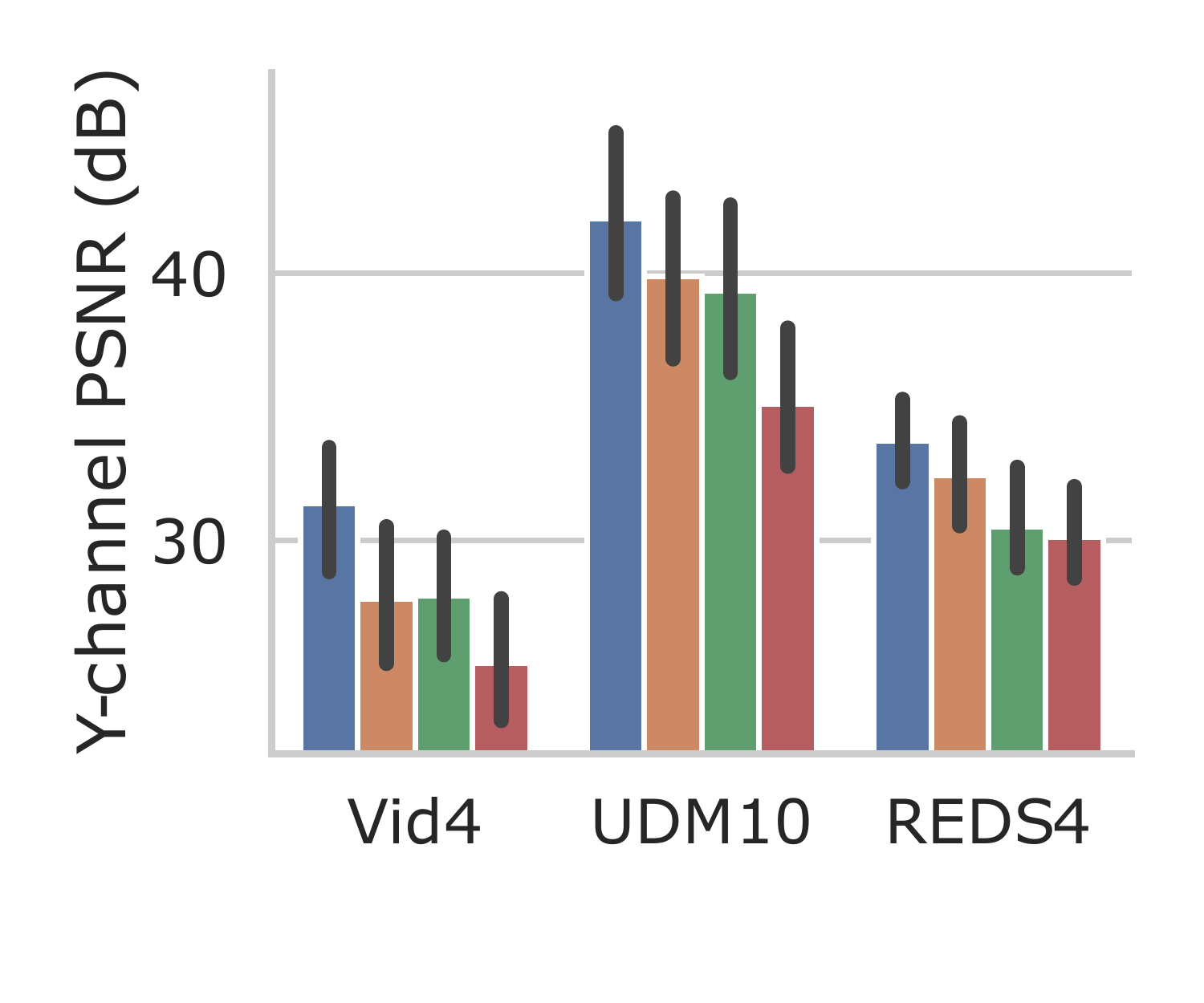}
    \end{subfigure}
    \begin{subfigure}[b]{0.59\linewidth}
        \centering
        \includegraphics[width=\linewidth]{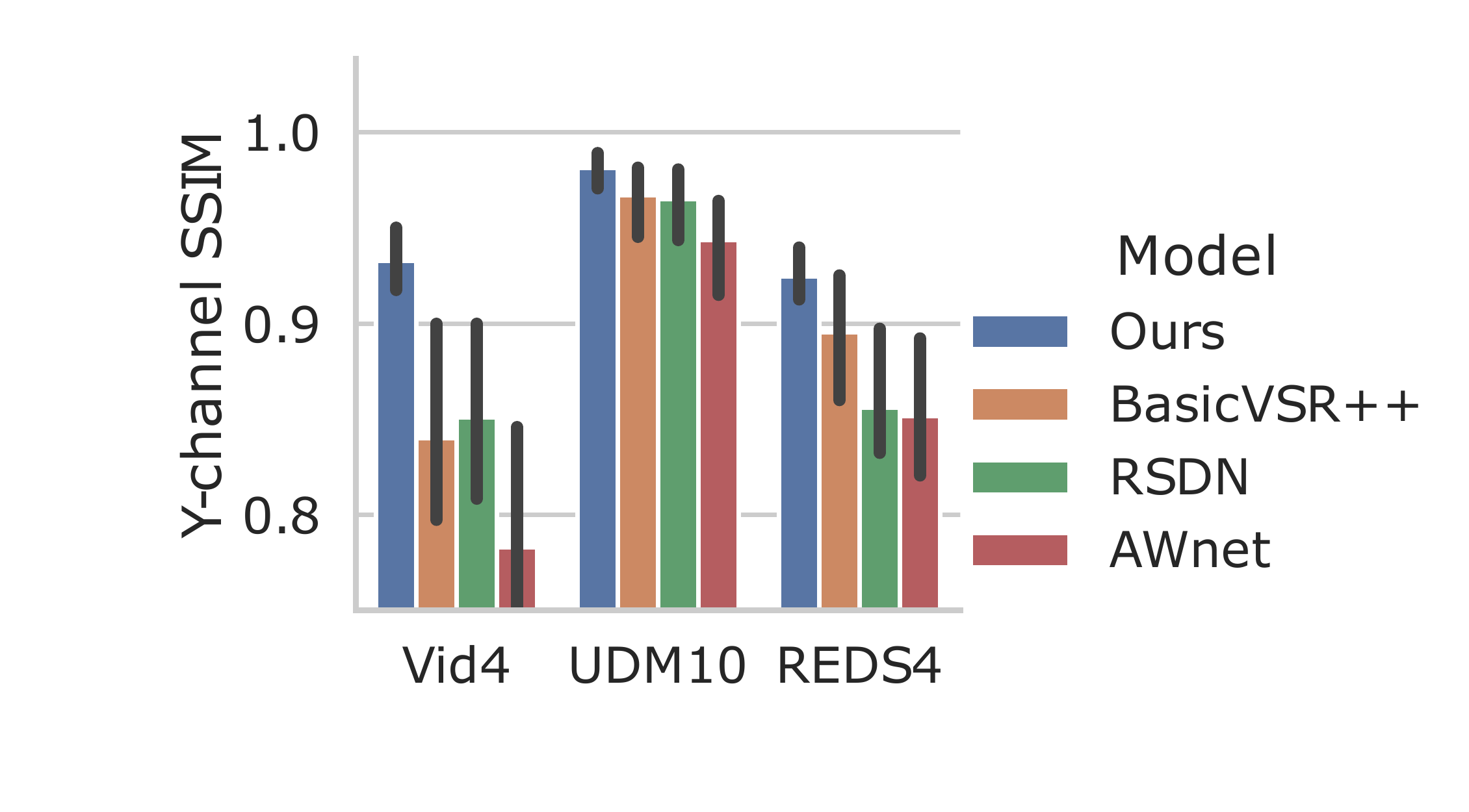}
    \end{subfigure}
    \vskip -0.25in
    \caption{Metrics are computed per-frame and averaged over all frames. Key-frames are excluded, for a fair comparison.}
    \vskip -0.1in
    \label{fig:vsr_comp}
\end{figure}

\begin{figure}[t]
    \centering
    \includegraphics[width=1.0\linewidth]{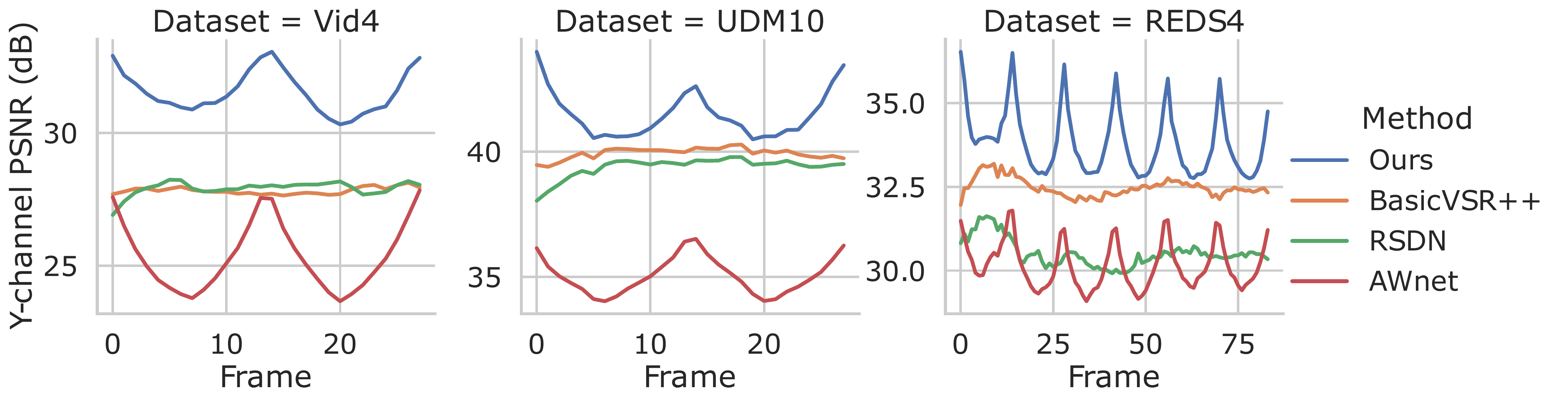}
    \vskip -0.15in
    \caption{{PSNR versus frame number.}}
    \vskip -0.18in
    \label{fig:psnr_prog}
\end{figure}

\vskip 0.05in\noindent{\bf Compared schemes.}
We benchmark our neural network on the standard datasets described above, by comparing them against the following baselines. As we describe in \xref{hardware_eval}, we also retrain these baselines with the data collected on our hardware for evaluating our prototype system.

\squishlist
    \item { {\it AWnet, BasicVSR++ and RSDN.}
    We first evaluate the quality of the image details generated by our method by comparing with the state-of-the-art dual camera as well as single camera super-resolution methods, AWnet \cite{Cheng2021ADC}, BasicVSR++ \cite{chan2021basicvsr} and RSDN \cite{10.1007/978-3-030-58610-2_38} that improve the resolution by the same scale (4x) as our system. Since these methods do not perform colorization, we only use the Y-channel of the output frames for comparisons.
    }
    \item {\it DEVC.} To understand how effectively our model can propagate color from key-frames, we compare the colorization
    performance of our method with the state-of-the-art color propagation method DEVC \cite{10.1145/3197517.3201365}
    that propagate color from a reference color frame to a sequence of grayscale frames. We convert the outputs of
    our method as well as DEVC, to CIELAB color space and evaluate the performance on ab-channels, which contain all
    of the color information. DEVC is a strong colorization baseline due to these design choices: i) DEVC proposes a recurrent network that propagates colorization history through time, which enables it to maintain temporal consistency in the colorization output. ii) It uses a ‘Correspondence Subnet’ block that spatially aligns history to the current input, similar to the deformable alignment module we used in our recurrent elements. iii) The neural network is end-to-end trained with video datasets, employing a temporal consistency loss as one of its loss components, to achieve temporally coherent output. iv) Finally, DEVC proposes an explicit video color propagation mode, where the color is propagated from the first color frame to subsequent grayscale frames. We evaluate our method with this mode, where we use the closest key-frame as the reference, and its color is propagated to neighboring frames.
    \item {\it TTSR.} We compare our use of temporal information from multiple frames  to existing reference-based methods that are for images and do not leverage
    similarities across frames. Since RefSR methods are primarily trained to use images in the wild as a reference,
    they might not be well-tuned to super-resolve and colorize the sequence of frames in a video. So we retrain the RefSR
    method TTSR \cite{inproceedings} with the same dataset we use for a fair comparison.
\squishend

\begin{figure}[t!]
    \centering
    \begin{subfigure}[b]{\linewidth}
        \centering
        \includegraphics[width=\linewidth]{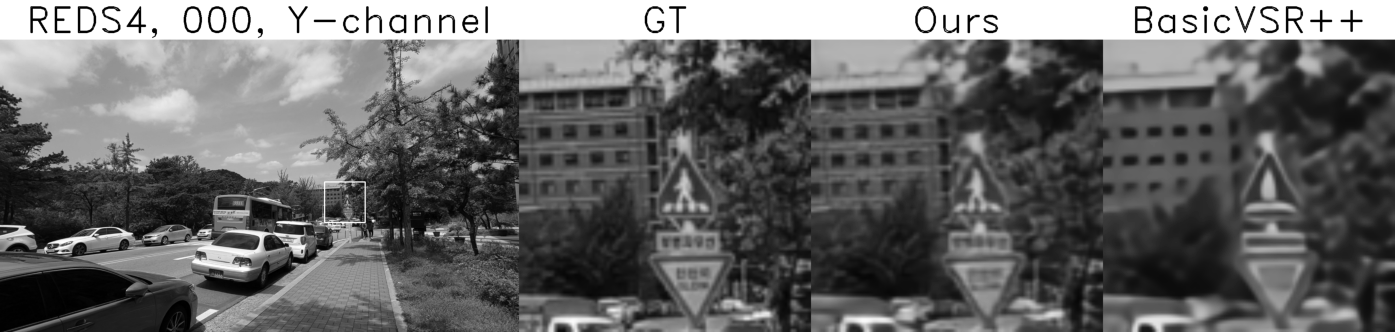}
    \end{subfigure}
    \begin{subfigure}[b]{\linewidth}
        \centering
        \includegraphics[width=\linewidth]{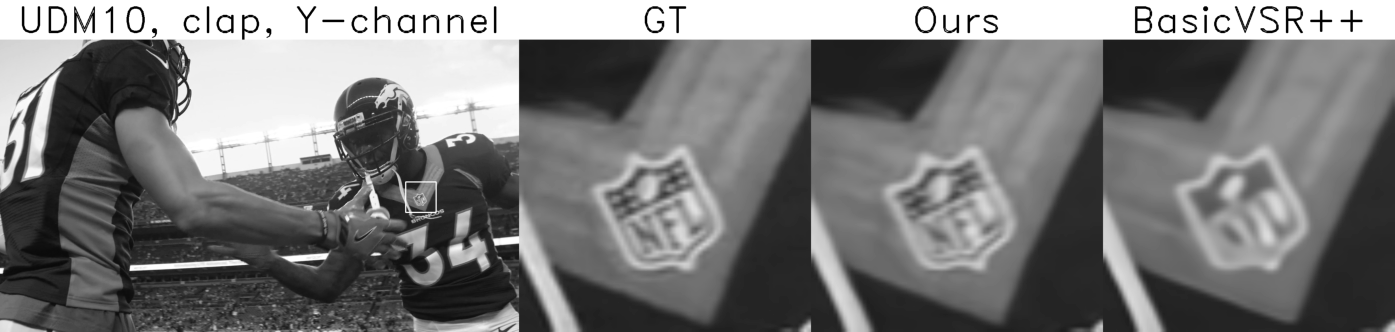}
    \end{subfigure}
    \begin{subfigure}[b]{\linewidth}
        \centering
        \includegraphics[width=\linewidth]{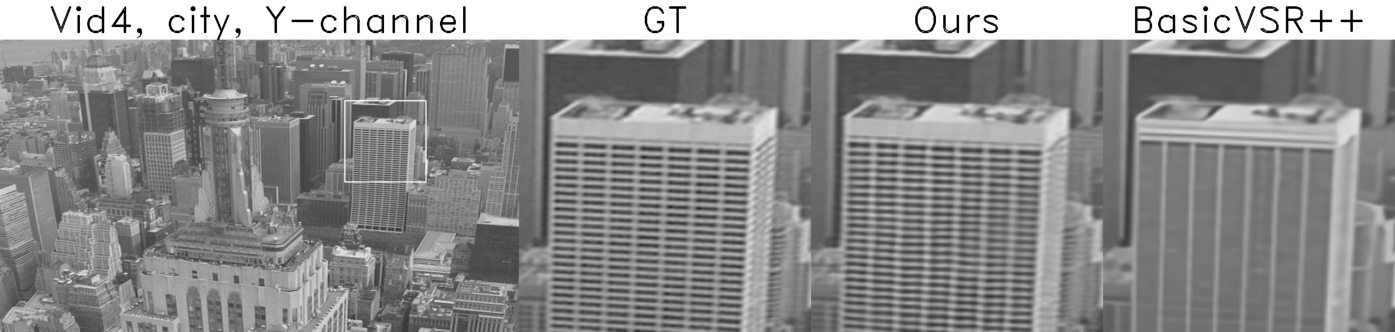}
    \end{subfigure}
    \vskip -0.12in
    \caption{Qualitative comparison of our key-frame approach with video super resolution methods. We compare the mid-point frame equidistant from both key frames, which is the worse case scenario for our system. GT is ground truth frame.}
    \vskip -0.15in
    \label{fig:vsr_qual_comp}
\end{figure}

\begin{figure}[t]
    \centering
    \begin{subfigure}[b]{0.43\linewidth}
        \centering
        \includegraphics[width=\linewidth]{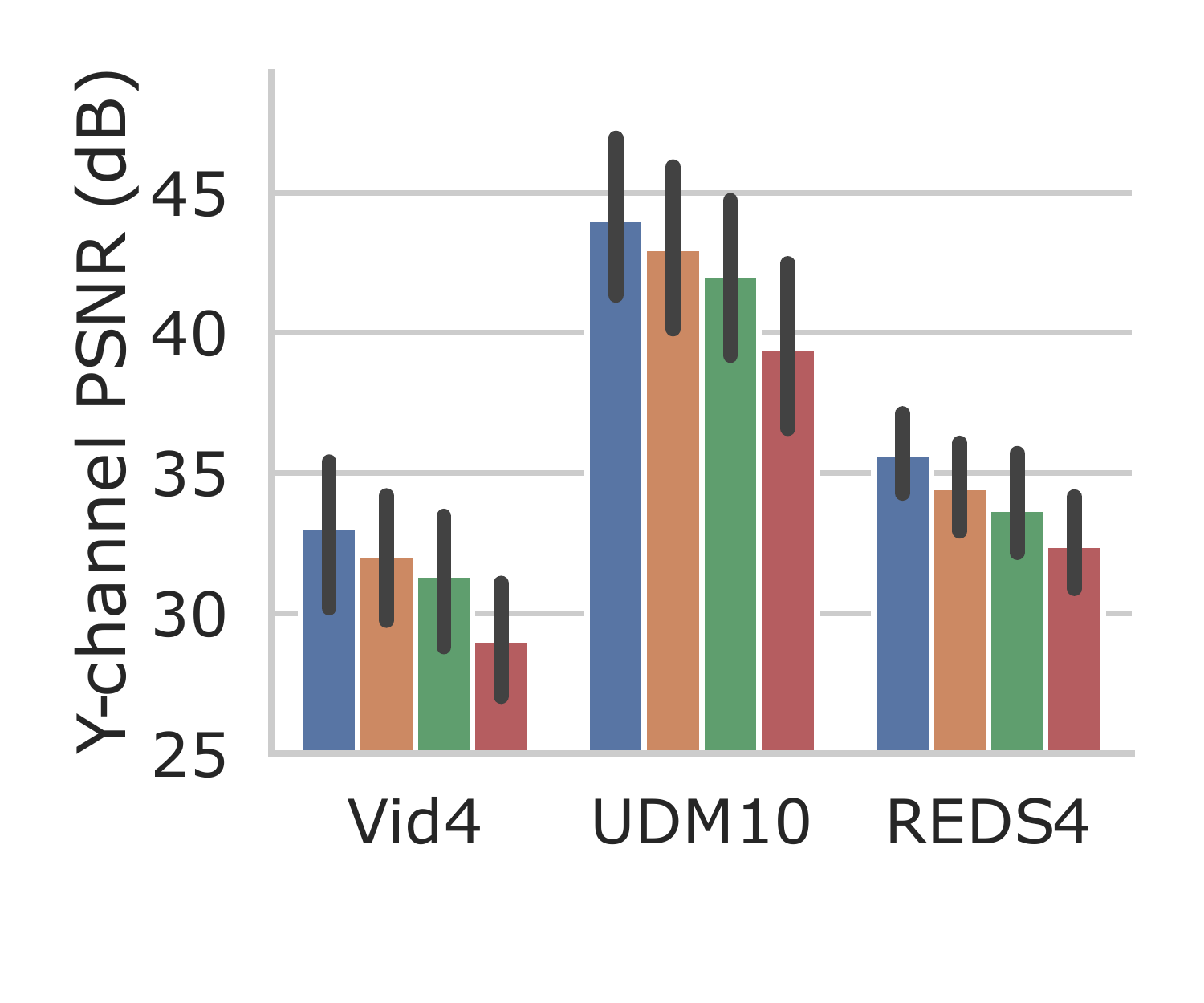}
    \end{subfigure}
    \begin{subfigure}[b]{0.55\linewidth}
        \centering
        \includegraphics[width=\linewidth]{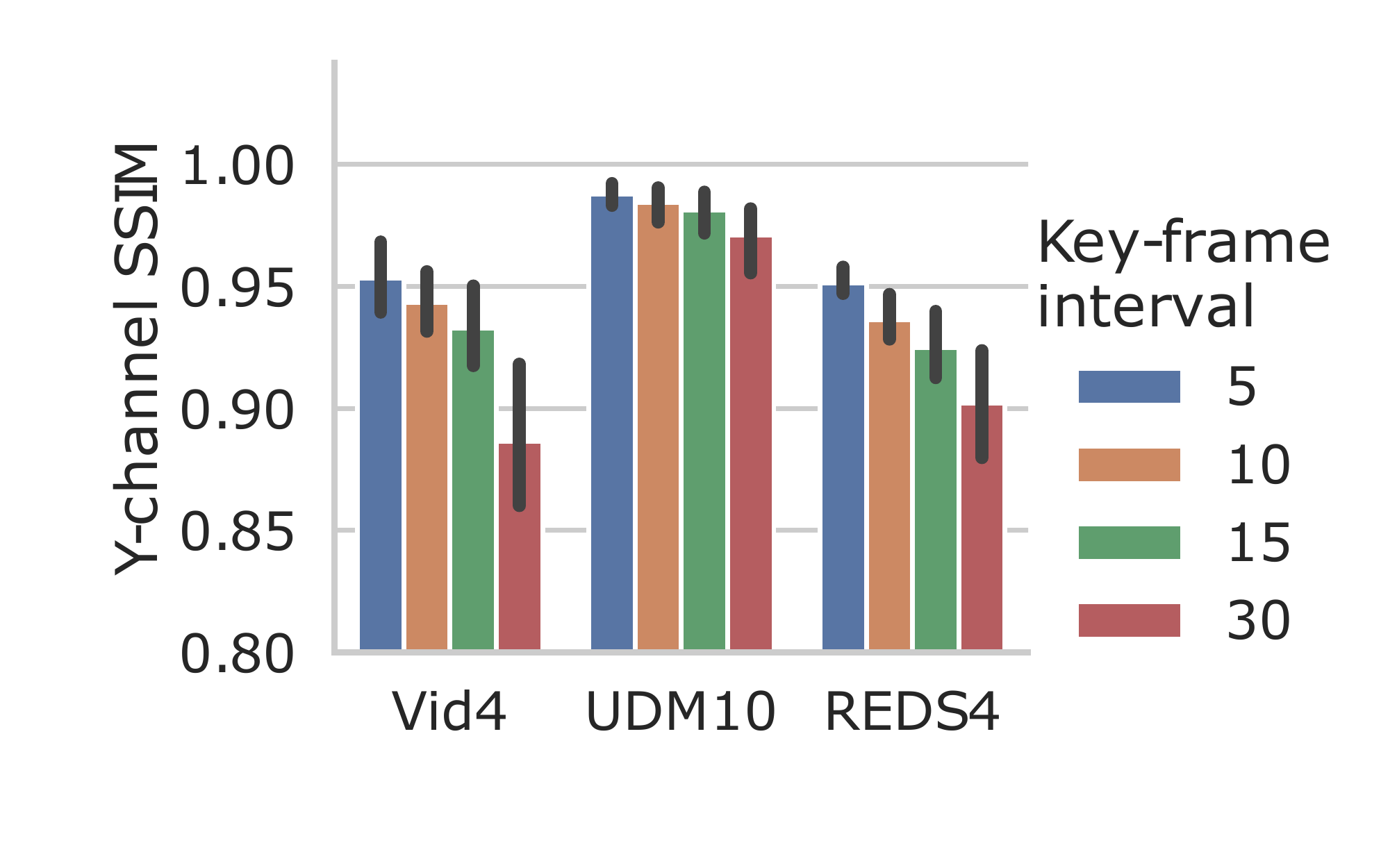}
    \end{subfigure}
    \vskip -0.25in
    \caption{{Comparison with different key-frame intervals.}}
    \vskip -0.2in
    \label{fig:key_frame_int_comp}
\end{figure}

{
\vskip 0.05in\noindent{\bf Evaluation of the quality of image details.} We evaluate the quality of image details both qualitatively and quantitatively, by comparing with AWnet, BasicVSR++ and RSDN. AWnet is a dual camera super-resolution method that super-resolves high frame-rate low-resolution video using low frame-rate high-resolution video. BasicVSR++ and RSDN are video super-resolution methods that convert a low-resolution video to a high-resolution video. Though these methods only perform the upscaling task and do not perform the colorization task, they act as strong baselines for image details in the output video stream. As a result, we  compare our Y-channel performance with that of AWnet, BasicVSR++ and RSDN.} 

{
Fig.~\ref{fig:vsr_comp} compares the Y-channel performance of our method with BasicVSR++ and RSDN. Each bar in the plot represents the distribution of average PSNR and SSIM over each video in the datasets. The results show a  1.7-4dB improvement in PSNR across the three datasets. While AWnet leverages the benefits of the dual cameras, it only uses previously generated output frame for recurrence. In contrast, our method, BasicVSR and RSDN are intrinsically recurrent networks that maintain a persistent hidden state to leverage temporal correlations between consecutive frames. Further, the improvements over BasicVSR++ and RSDN shows the effectiveness of key-frames in achieving improved image details. We also examine  the frame-by-frame PSNR progression and compare it to that of prior methods. Fig.~\ref{fig:psnr_prog} plots the PSNRs for each frame averaged across all videos in each dataset. The  key-frames improve the  quality of the output video, with larger gains closer to the key-frames.} 

{
In Fig.~\ref{fig:vsr_qual_comp}, we compare frame 7 of a video from each of the datasets. Since the key-frame interval is 15, frame 7 is farthest from forward and backward key-frame and hence serves as the worst case scenario for our  model. The plots show that our method is able to more faithfully reconstruct the fine details such as signs, letters, shapes and textures.} 

\begin{figure}[t]
    \centering
    \begin{subfigure}[b]{0.4\linewidth}
        \centering
        \includegraphics[width=\linewidth]{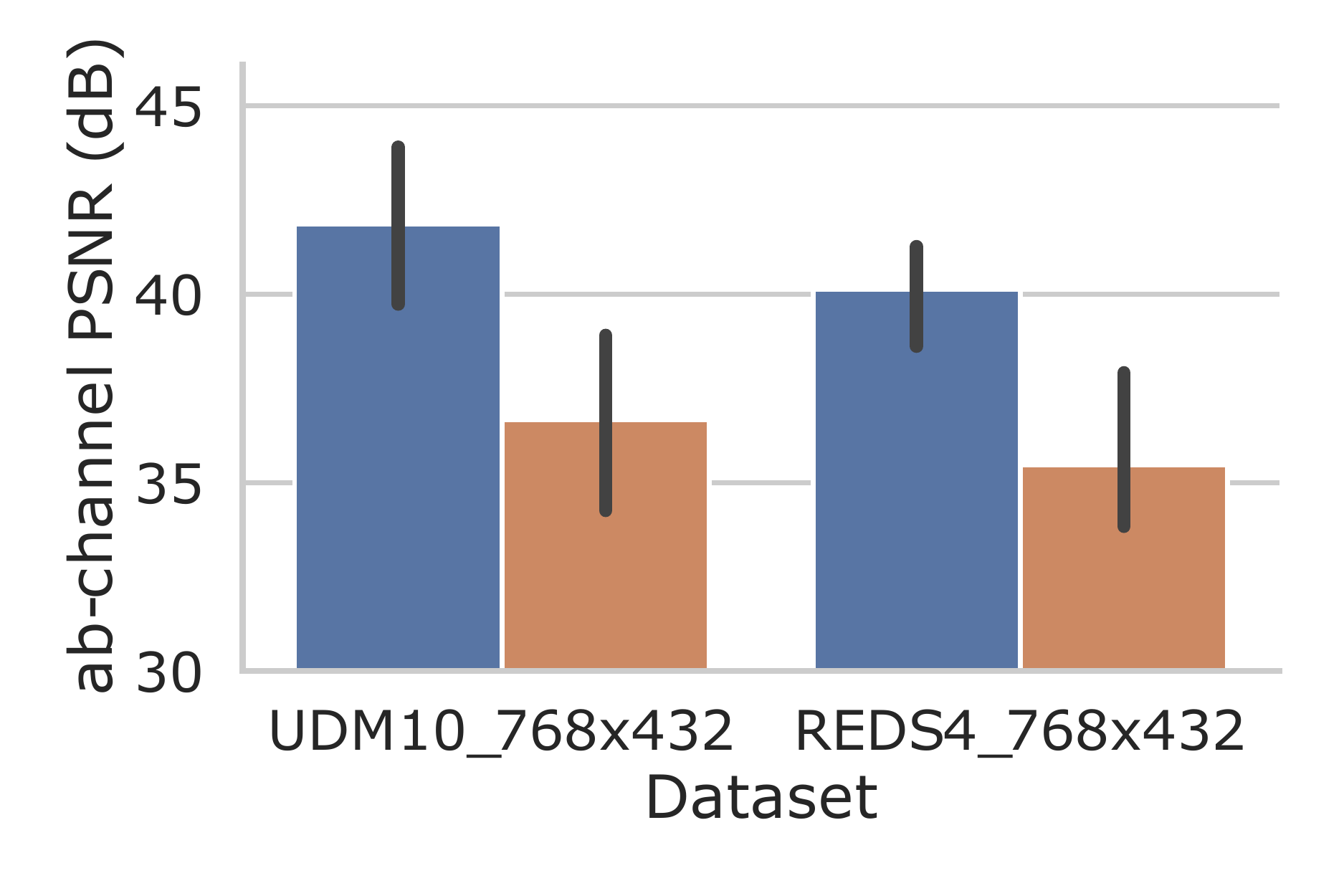}
    \end{subfigure}
    \begin{subfigure}[b]{0.5\linewidth}
        \centering
        \includegraphics[width=\linewidth]{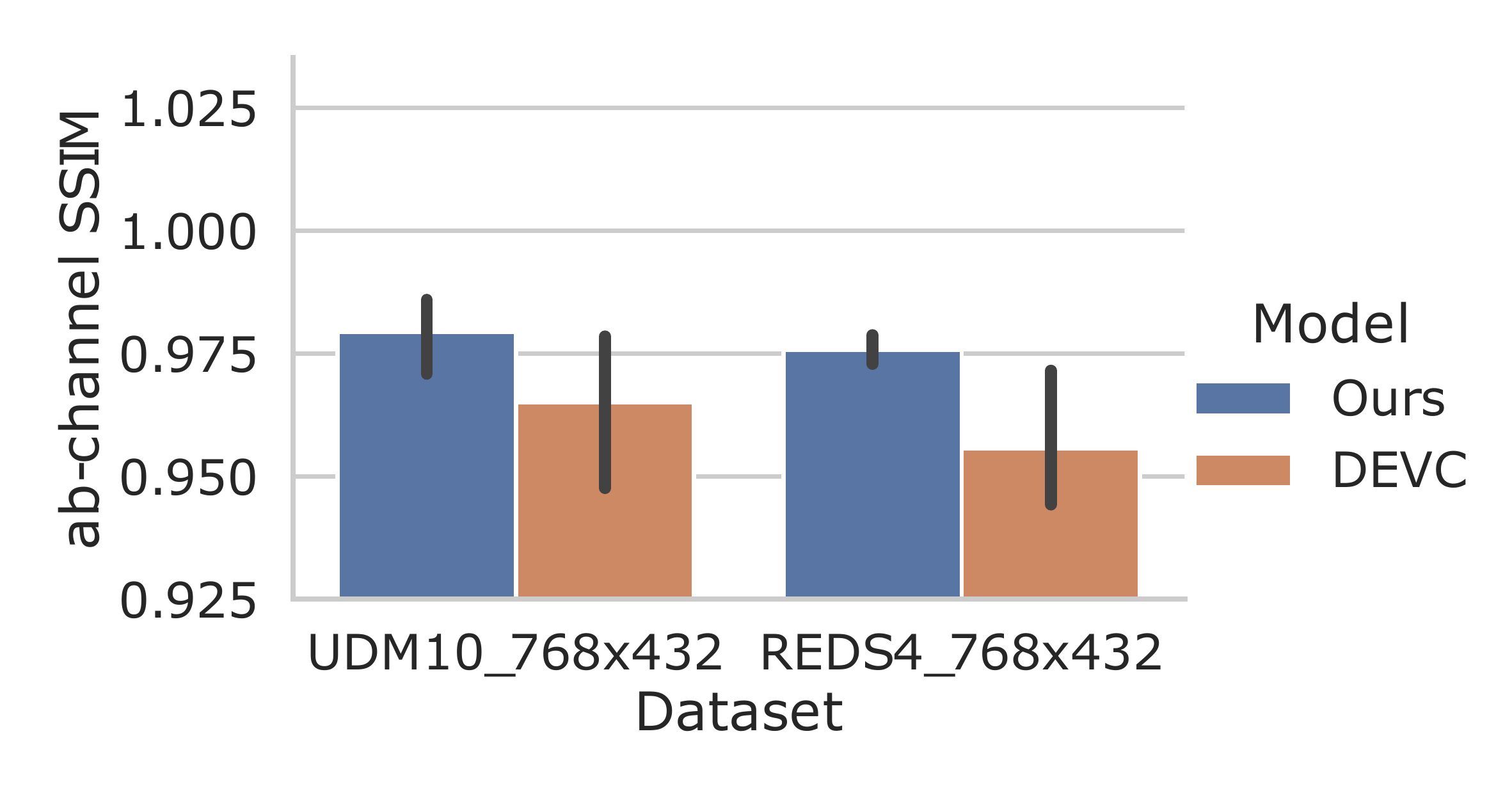}
    \end{subfigure}
    \vskip -0.2in
    \caption{We compare  with DEVC which does not perform super resolution but can propagate color.}
    \vskip -0.15in
    \label{fig:colorprop_comp}
\end{figure}

\vskip 0.05in\noindent{\bf Effectiveness of color propagation.}
Obtaining color is an additional challenge in low-power wireless video  as color sensors are more power hungry
compared to monochrome sensors. To mitigate this problem, in our method, we use low-resolution grayscale sensor
as our primary sensor and use key-frames captured from the high-resolution color sensor to colorize the output.
While our method upscales the input to higher resolution in addition to colorization, the colorization task is 
similar to deep learning methods proposed for color propagation which  colorize a grayscale
video using a color frame as reference. To evaluate the colorization aspect of our model, we compare our model's color output  with the  state-of-the-art color propagation
method DEVC \cite{10.1145/3197517.3201365}.

DEVC natively operates on 768x432 resolution, and as result we center crop the high-resolution test sets UDM10
and REDS4 to 768x384 to obtain UDM10\_768x432 and REDS4\_768 x432 datasets. We evaluate  DEVC  using frame 0 as the reference, propagating color for one key-frame interval i.e., 15 frames.
To compare the colorization performance, we convert the outputs to CIELAB colorspace and extract the ab-channels,
which comprise of all the color information. Fig.~\ref{fig:colorprop_comp} shows the average PSNRs and SSIMs for  both UDM10\_768x432 and REDS4\_768x432 datasets showing a 4--6~dB gain. 

\begin{figure}
    \centering
    \begin{subfigure}[b]{\linewidth}
        \centering
        \includegraphics[width=\linewidth]{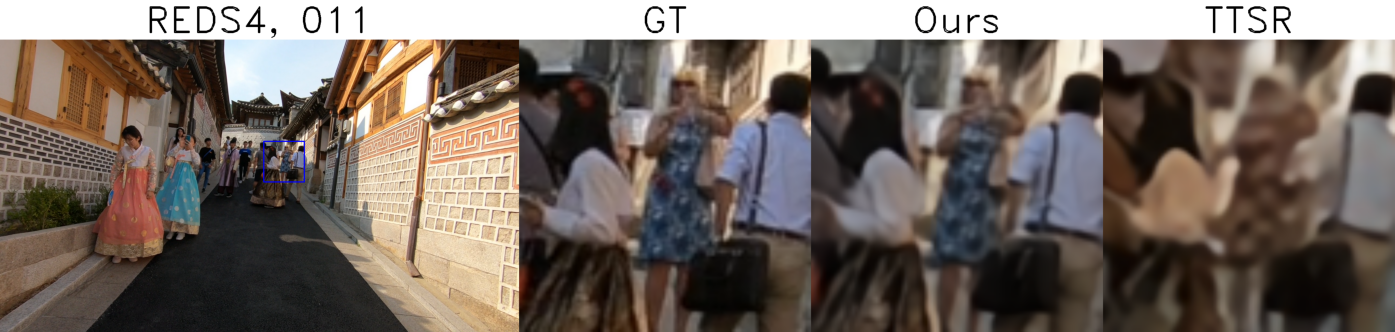}
    \end{subfigure}
    \begin{subfigure}[b]{\linewidth}
        \centering
        \includegraphics[width=\linewidth]{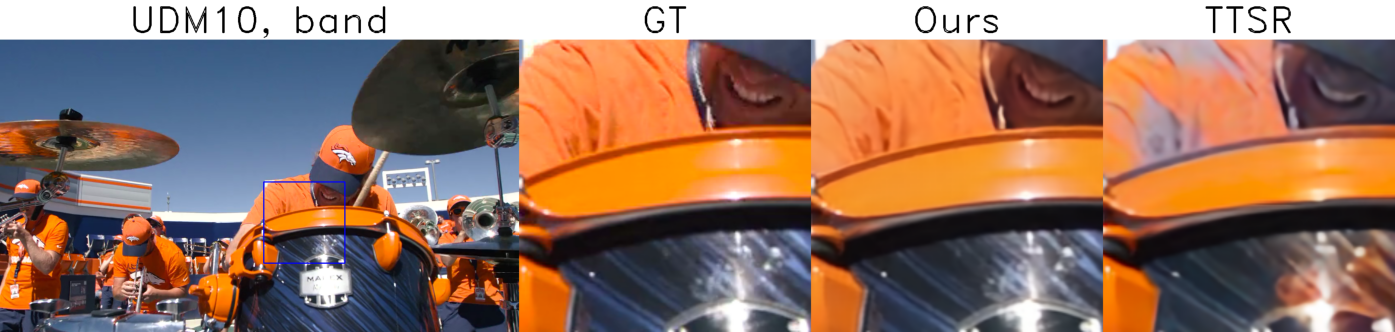}
    \end{subfigure}
    \begin{subfigure}[b]{\linewidth}
        \centering
        \includegraphics[width=\linewidth]{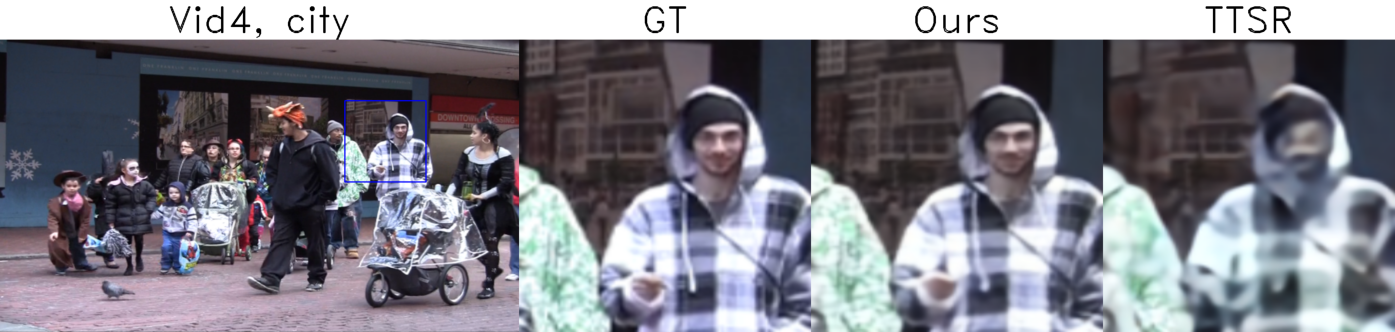}
    \end{subfigure}
    \vskip -0.1in
    \caption{Qualitative comparisons with reference-based {\it image} super-resolution methods.}
    \vskip -0.15in
    \label{fig:refsr_qual_comp}
\end{figure}

\vskip 0.05in\noindent{\bf Aggregating multi-frame information.} Existing methods that use reference images to super-resolve low-resolution images, similar to our use of
key-frames, are \textit{reference based image super-resolution} methods (RefSR). Since the
reference could just be a key-frame, the architecture is suitable for key-frame wireless video  applications.
However, RefSR methods have only been proposed for images and as a result do not leverage multi-frame
information that methods designed for video do. To evaluate the benefits of the bidirectional propagation
of our method, we compare with the RefSR method TTSR \cite{inproceedings}. TTSR however, is not originally  trained on
video datasets and as a result would have poor performance when applied to video. So, we retrained TTSR
 on Vimeo-90K from scratch using nearest key-frames as references, and use the resulting model for
 comparison. Since TTSR can use color key-frame as the reference, the model can be trained to colorize
input low-resolution frame, and so  we could use the full RGB-channel performance in this case.

\begin{figure}
    \centering
    \begin{subfigure}[b]{0.4\linewidth}
        \centering
        \includegraphics[width=\linewidth]{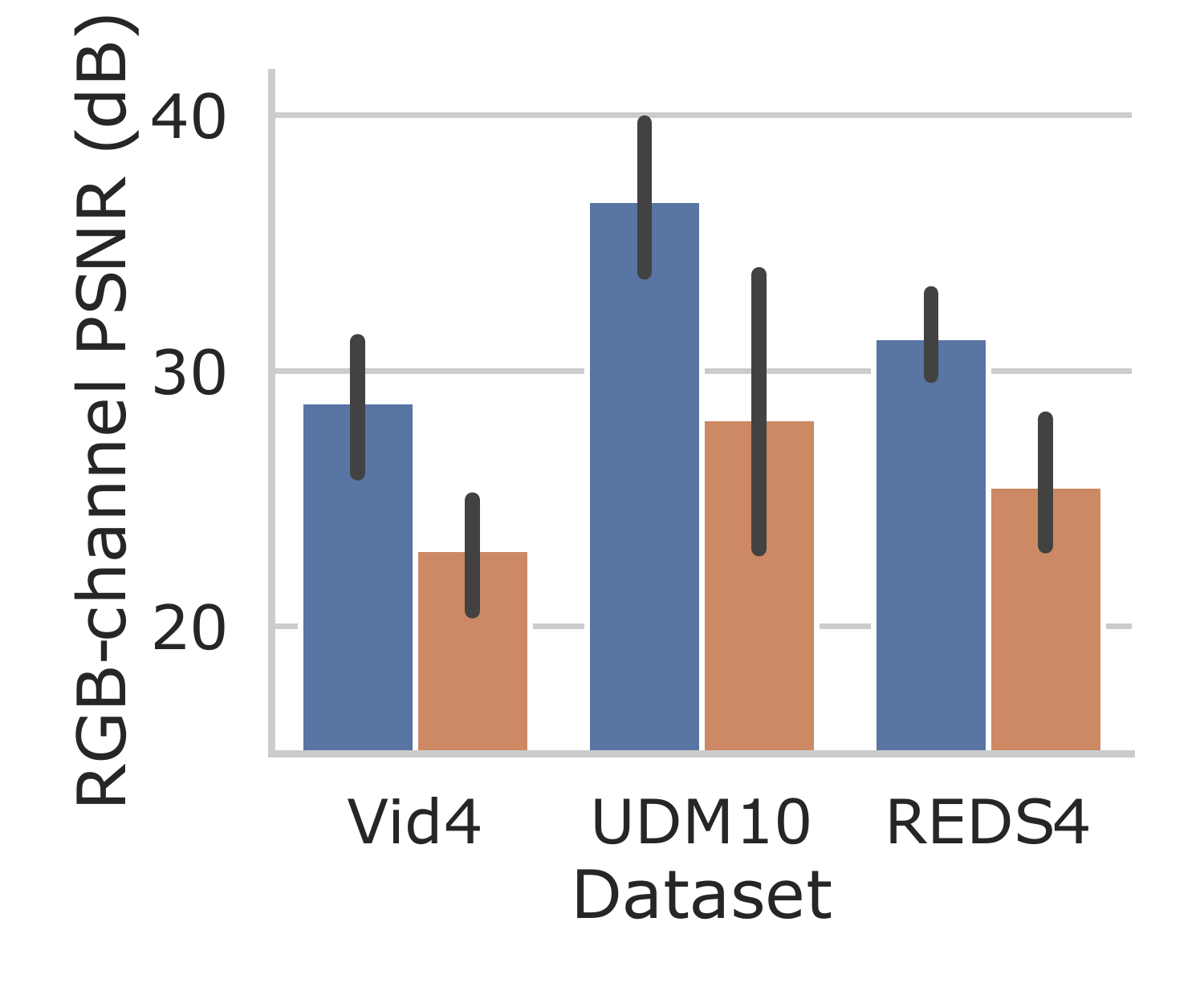}
    \end{subfigure}
    \begin{subfigure}[b]{0.5\linewidth}
        \centering
        \includegraphics[width=\linewidth]{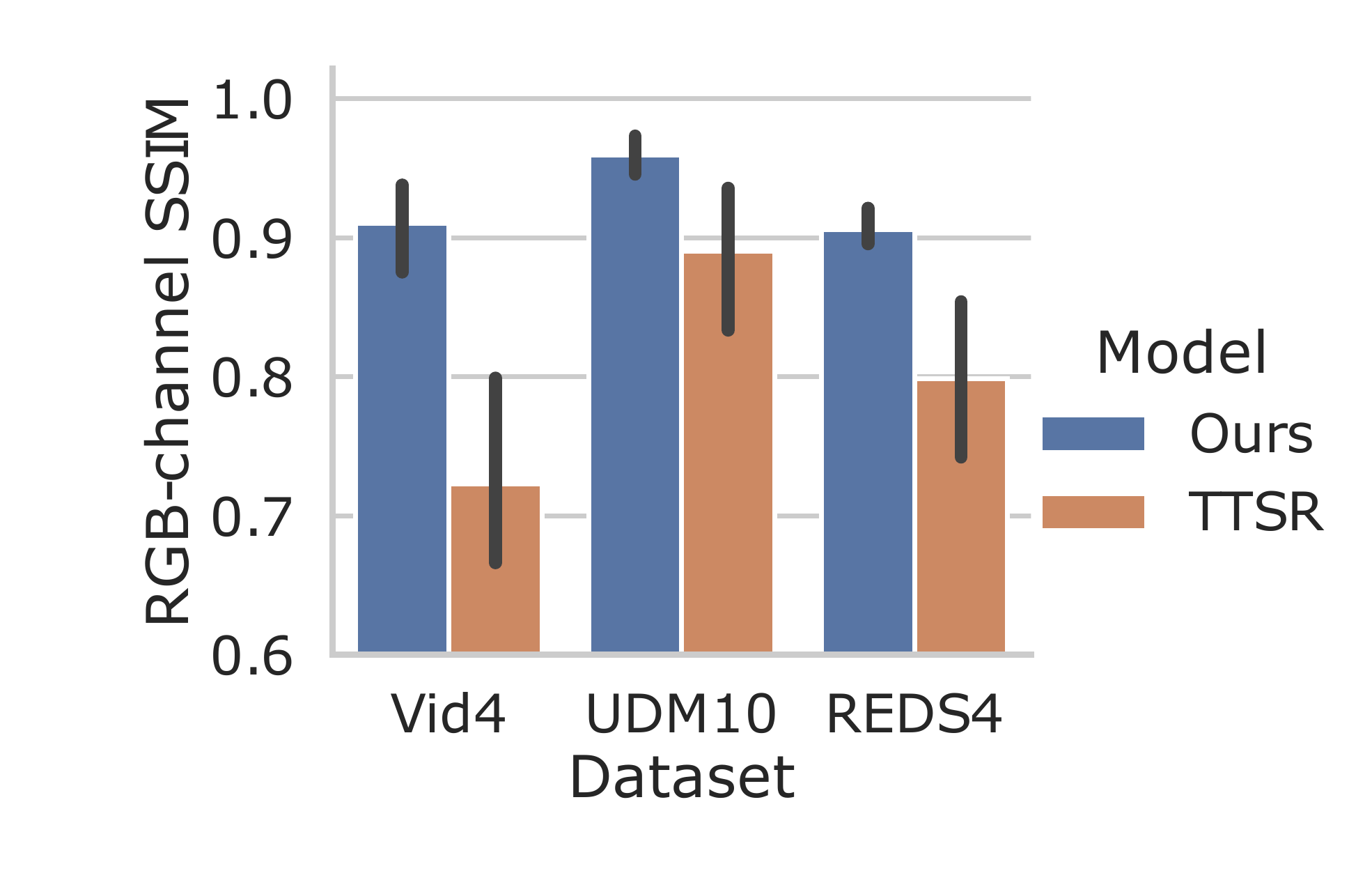}
    \end{subfigure}
    \vskip -0.15in
    \caption{Quantitative comparisons with reference-based {\it image} super-resolution methods.}
    \label{fig:refsr_comp}
\end{figure}

We compare our method and TTSR in Figs.~\ref{fig:refsr_qual_comp}
and \ref{fig:refsr_comp}. Frame 7 of
a video from one of the video in each dataset is compared as it is farthest from forward and backward
key-frame and hence serves as the worst case scenario for our model. Our method is able to propagate the color
and details from key-frames more effectively than TTSR, because our method aggregates the information from
neighbouring low-resolution frames, as well as the forward and backward key-frames in a recurrent fashion. Fig.~\ref{fig:refsr_comp} shows that on average we observe a gain of 4-7~dB.

\begin{figure}
    \centering
    \vskip -0.19in
    \begin{subfigure}[b]{\linewidth}
        \centering
        \includegraphics[width=0.95\linewidth]{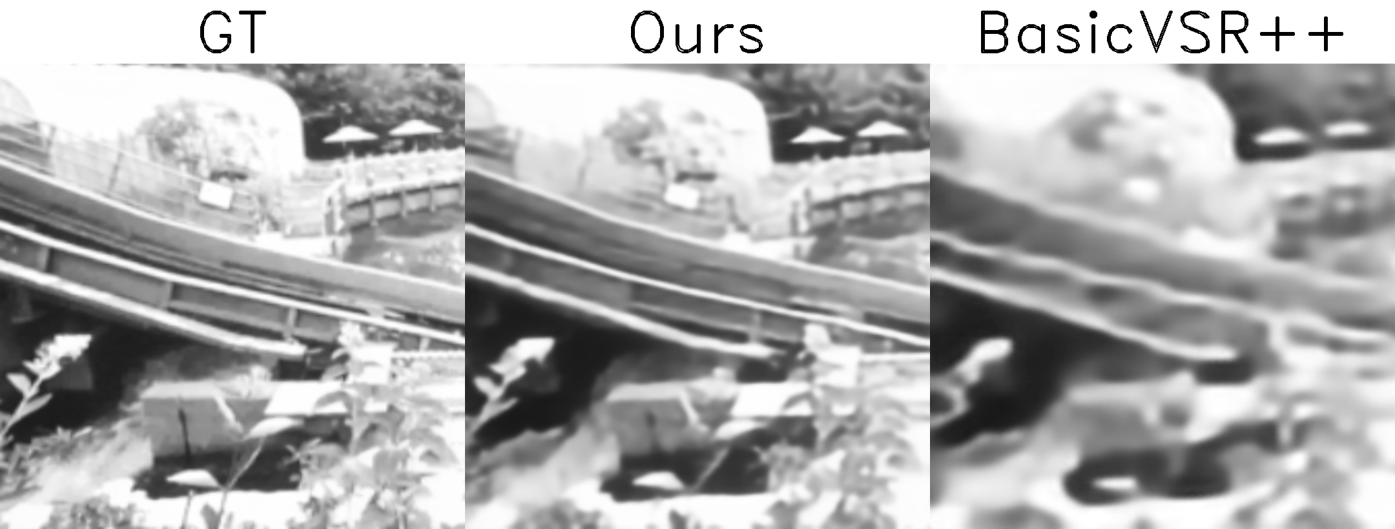}
        \caption{Y-channel comparison with video super-resolution.}
    \end{subfigure}
    \begin{subfigure}[b]{\linewidth}
        \centering
        \includegraphics[width=0.95\linewidth]{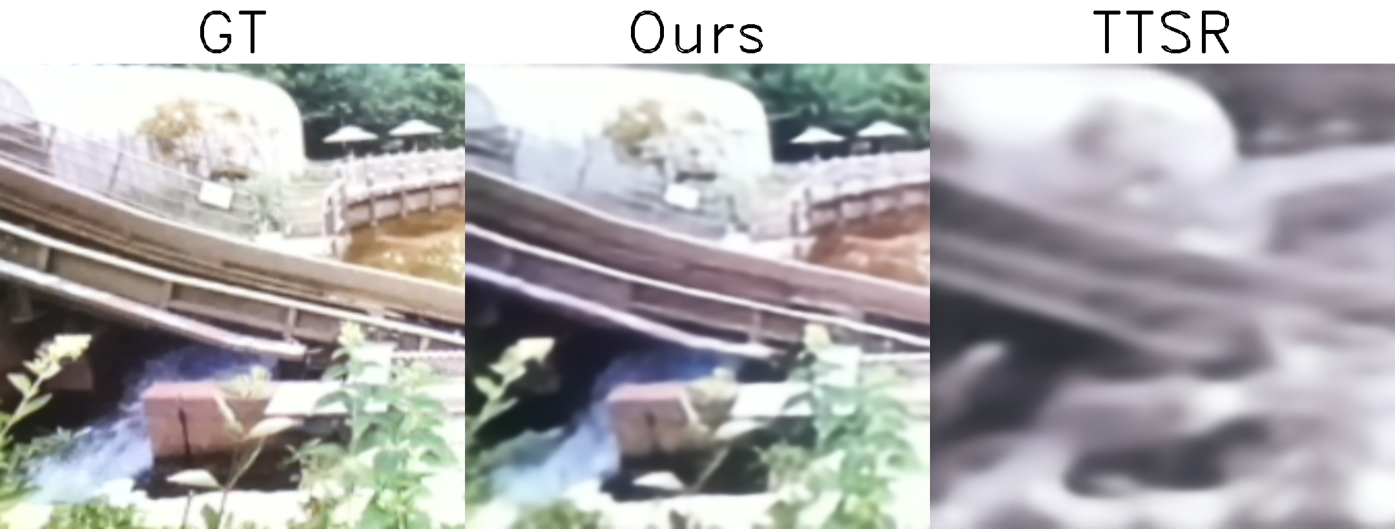}
        \caption{RGB-channel comparison with Reference-based method.}
    \end{subfigure}
    \vskip -0.1in
    \caption{Qualitative comparison on real world data.}
    \vskip -0.15in
    \label{fig:3c_qual_comp}
\end{figure}

\subsection{Hardware evaluation}
\label{hardware_eval}


The real-world resolution degradation process is much different from the standardized bicubic downsampling process and moreover, varies from one sensor to the other. Using a second camera for capturing key-frames results in image and color artifacts induced by the sensor. In addition to the sensor characteristics, low-resolution and high-resolution streams have different perspectives due to a small difference in their spatial locations. To address these concerns, we evaluate the system's performance in presence of these real-world effects.

\vskip 0.05in\noindent{\bf Capturing the ground truth data.} We use a low-power micro-controller for our hardware prototype, so it 
only supports high-resolution color video at 1 fps. Training the network, however, requires 15 fps ground truth video
stream that has a one-to-one correspondence to the 15fps low resolution stream. To overcome this problem, we use a third Raspberry Pi camera~\cite{picamera2} that can operate at VGA resolution to collect the ground-truth data.  For effective training, the input streams have to be synchronized with
the ground truth stream, which can be challenging. We address the synchronization problem by measuring the relative timestamps. At the start of the data collection, the Raspberry Pi initiates a local counter, and at the same time, sends a synchronization pulse to both low-resolution and high-resolution
microcontrollers. We observed this method achieves 3 synchronized timers for each of the 3 video streams, separated by less than 10ms,
which is sufficient to synchronize  15fps video streams.


\vskip 0.05in\noindent{\bf Training and evaluation procedure.}
 To evaluate   against a similar level of high mobility as is seen in the first-person point-of-view REDS dataset, we  capture the entire
REDS dataset \cite{Nah_2019_CVPR_Workshops_REDS} projected on a screen using our three camera setup. With this setup, we capture
a total of 11565 ground truth and low-resolution frames, and, 877 key-frames. We then split the dataset into 224 samples, with
each sample approximately containing 50-60 frames, without   jump cuts. The 224 samples are then split into non-overlapping train and evaluation sets containing 220 and 4 samples, respectively. Since the
resulting dataset is  smaller than typical simulated datasets, in addition to all the augmentations we apply, in every
epoch, we train on each video sample multiple times starting at a random start frame. Using these techniques, we retrain our method
as well are BasicVSR++ and TTSR for  40k forward and backward passes with a batch size of 8. 

\begin{figure}
    \centering
    \begin{subfigure}[b]{0.35\linewidth}
        \centering
        \includegraphics[width=\linewidth]{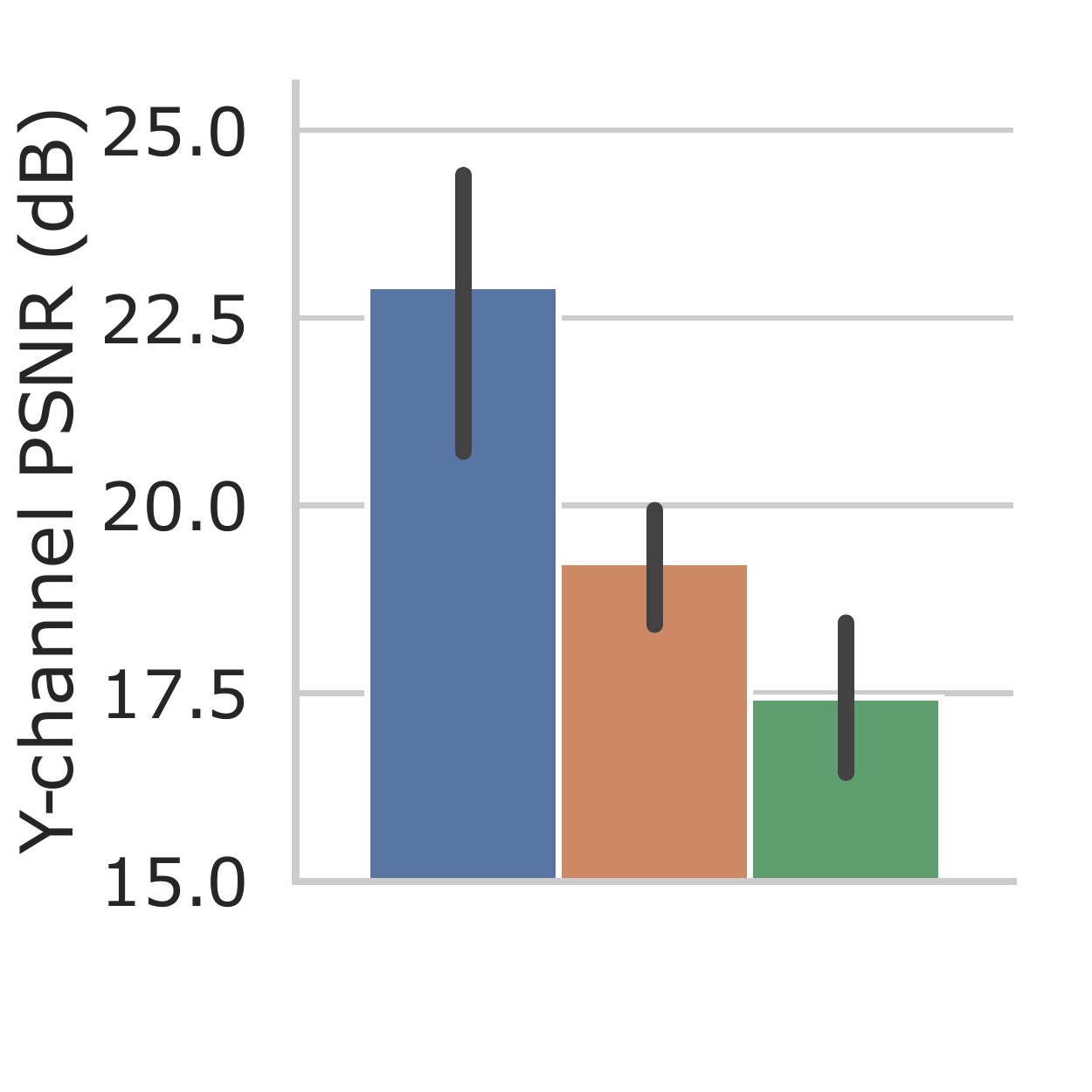}
    \end{subfigure}
    \begin{subfigure}[b]{0.6\linewidth}
        \centering
        \includegraphics[width=\linewidth]{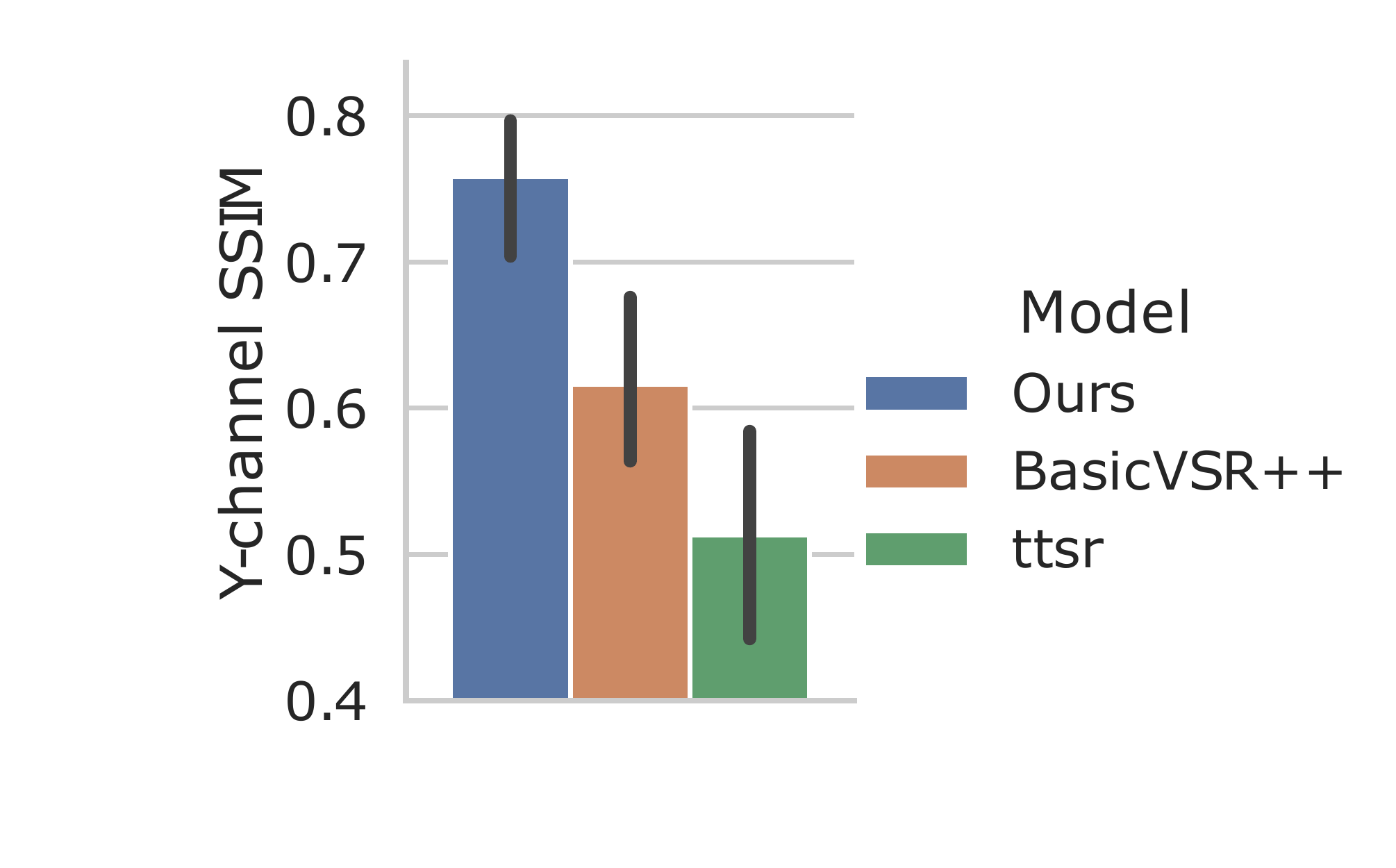}
    \end{subfigure}
    \vskip -0.2in
    \caption{Quantitative comparison of Y-channel video quality using data from our prototype  hardware.}
    \vskip -0.15in
    \label{fig:3c_comp_y}
\end{figure}

\vskip 0.05in\noindent{\bf Evaluation with hardware data.}  Fig.~\ref{fig:3c_comp_y} shows an  average gain of 3.7dB in Y-channel performance compared to BasicVSR++, more than the 1.7dB gain we observed for REDS4 dataset with simulated downsampling. This shows that our key-frame
approach adapts well to the real-world data. The improvement in performance gain shows that our method performs better
when the amount of training data is limited. Fig.~\ref{fig:3c_qual_comp} also shows that relative to the ground truth, our model generates much
clearer details compared to BasicVSR++.  The hardware results  show the overall effectiveness of key-frames in case of limited data. 
 In Fig.~\ref{fig:3c_comp_rgb}, we observe an RGB performance gain of 5.6 dB over TTSR, correlating well with the 5.7d B gain  on the standard datasets. Fig.~\ref{fig:3c_qual_comp} shows that
our bidirectional recurrent model with attention feature filters, leverages temporal information to  better map  color
and high-frequency details from key-frames.


\begin{figure}
    \centering
    \begin{subfigure}[b]{0.4\linewidth}
        \centering
        \includegraphics[width=\linewidth]{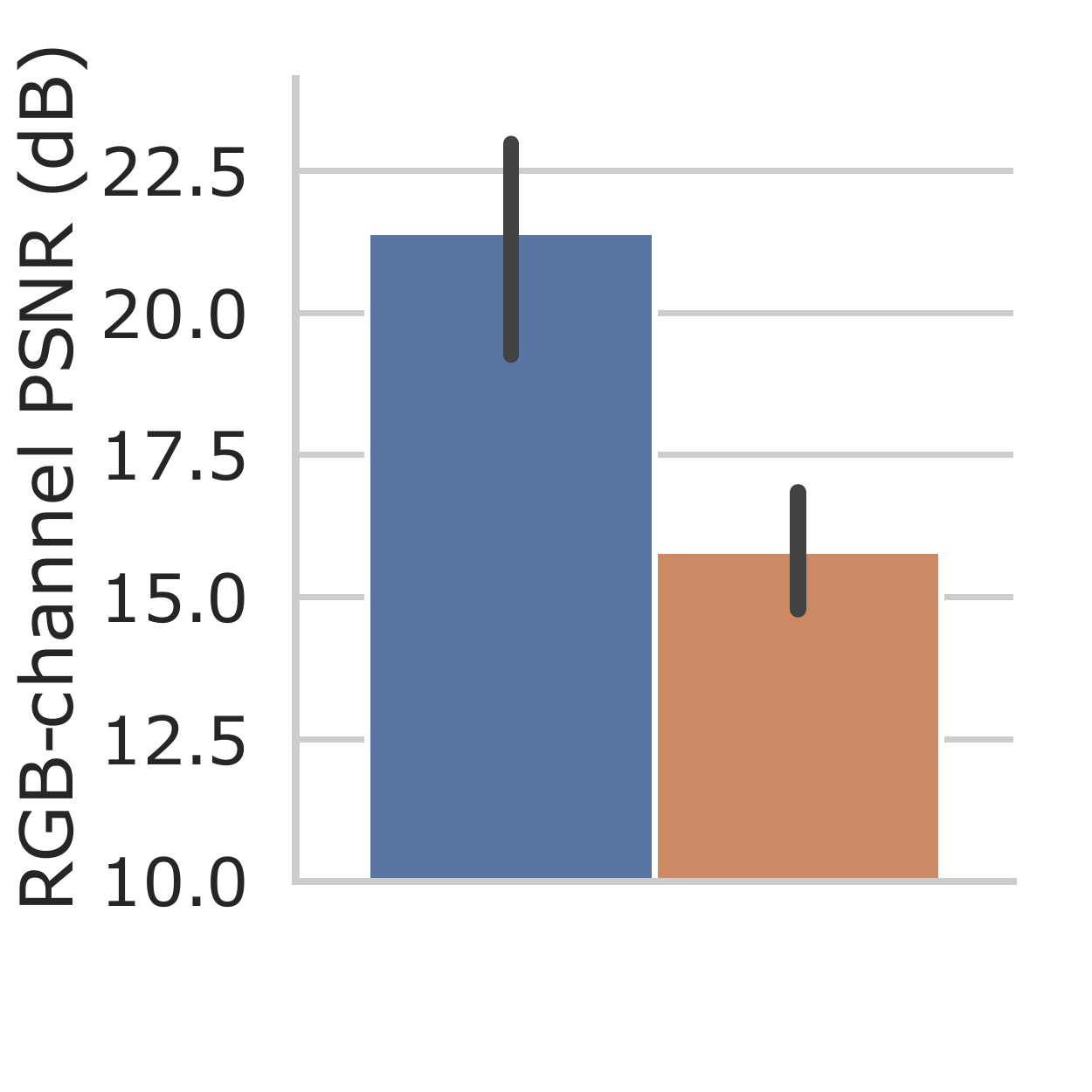}
    \end{subfigure}
    \begin{subfigure}[b]{0.55\linewidth}
        \centering
        \includegraphics[width=\linewidth]{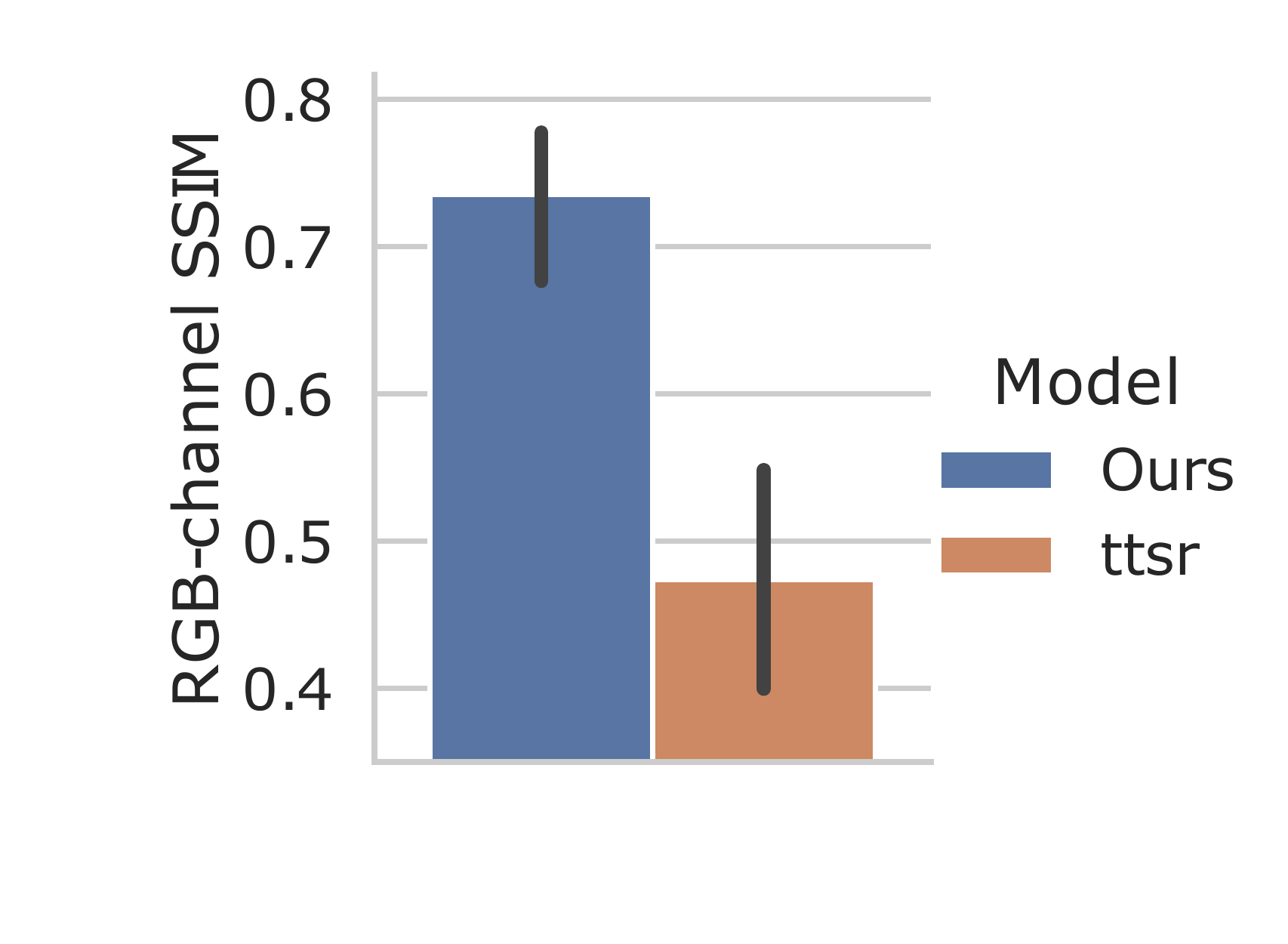}
    \end{subfigure}
    \vskip -0.25in
    \caption{Quantitative comparison of RGB-channel video quality using data from our prototype  hardware.}
    \vskip -0.1in
    \label{fig:3c_comp_rgb}
\end{figure}

\begin{figure}
        \begin{subfigure}[b]{0.35\linewidth}
        \centering
        \includegraphics[width=\linewidth]{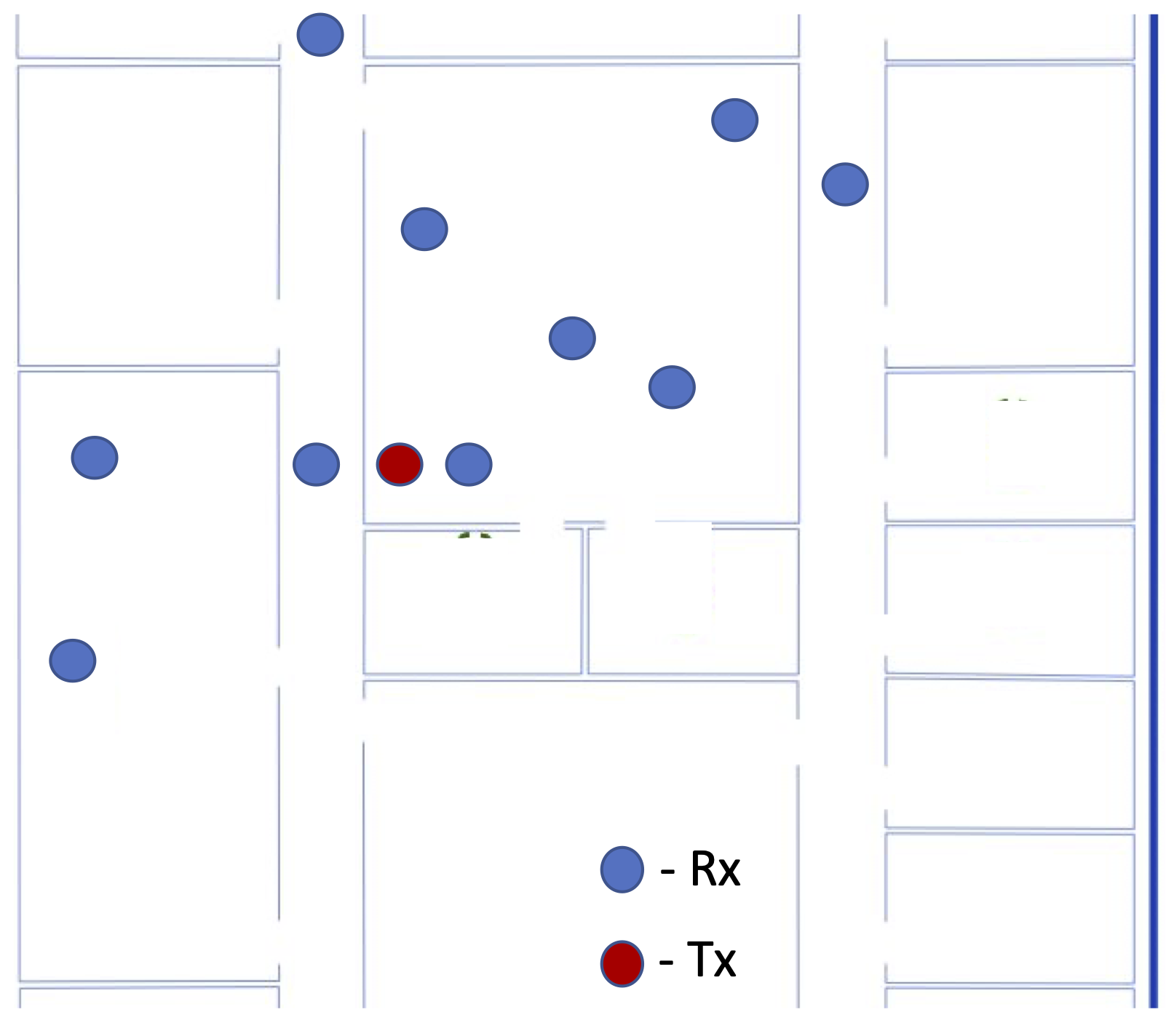}
         \caption{Map of camera locations.}
    \end{subfigure}
    \hspace{1em}
  \begin{subfigure}[b]{0.5\linewidth}
        \centering
        \includegraphics[width=\linewidth]{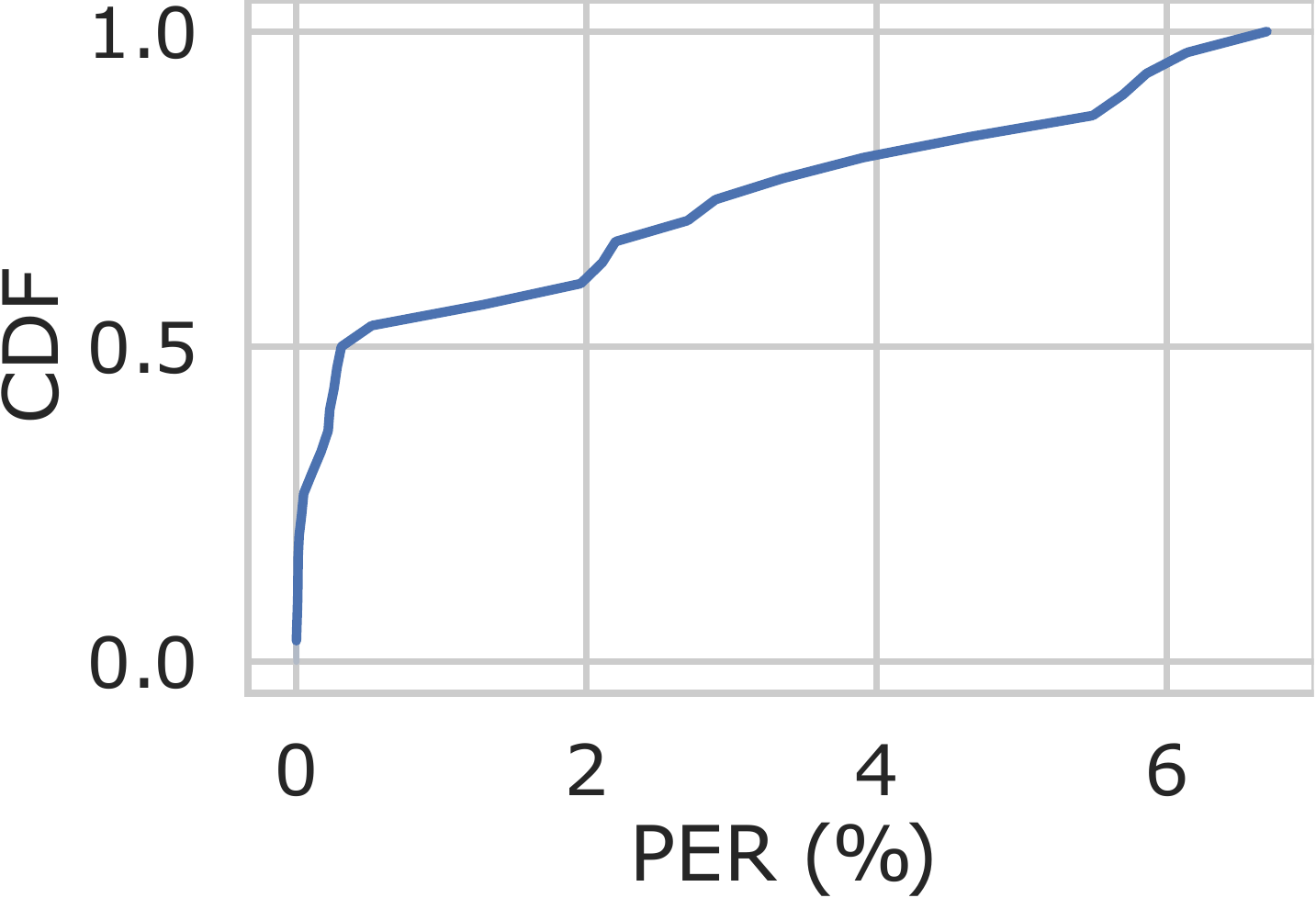}
        \caption{CDF of packet error rate.}
    \end{subfigure}
\vskip -0.1in
    \caption{Measured PERs across the testbed.}
    \vskip -0.15in
    \label{fig:per}
\end{figure}

\vskip 0.05in\noindent{\bf Addressing wireless packet losses.}  We use the testbed in Fig.~\ref{fig:per}a 
 to understand the packet-error rate in our system. We fix the transmitter location  and place the receiver across 10  locations.  The tested locations span a variety of line-of-sight, non-line-of-sight as well as locations that are not in the same room. At each receiver location, we transmit 1000 packets from the transmitter and compute the packet error rate (PER) at the receiver. We repeat this three times at each location for  a total of 30,000 transmitted packets.   Fig.~\ref{fig:per}b plots the CDF of the PER across all the receiver locations. The plot shows that the worse case PER is around 7\%. Note that since the packets come with a CRC, we know which of the packets have been incorrectly received. So an incorrect packet can be considered as missing data in the corresponding video frames.  Fig.~\ref{fig:packet_loss} shows the bicubic interpolation method we implemented to correct the lines lost due to packet error. For most cases, we observe that no noticeable artifacts remain in the corrected images. Even in adverse scenarios where multiple line losses occur together and interpolation method does not have the required information (e.g., last sample of Fig.~\ref{fig:packet_loss}), the artifacts are much less pronounced. 

\begin{figure}[t]
    \centering
    \begin{subfigure}[b]{0.9\linewidth}
        \centering
        \includegraphics[width=\linewidth]{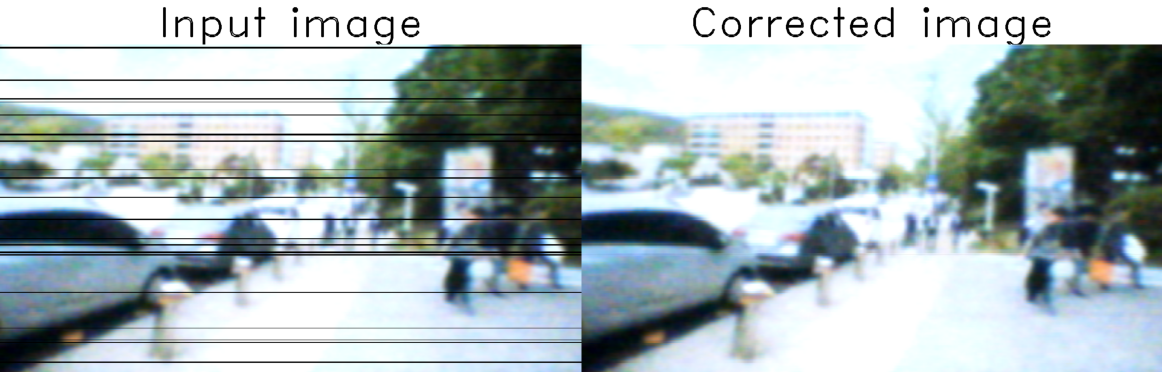}
    \end{subfigure}
    \begin{subfigure}[b]{0.9\linewidth}
        \centering
        \includegraphics[width=\linewidth]{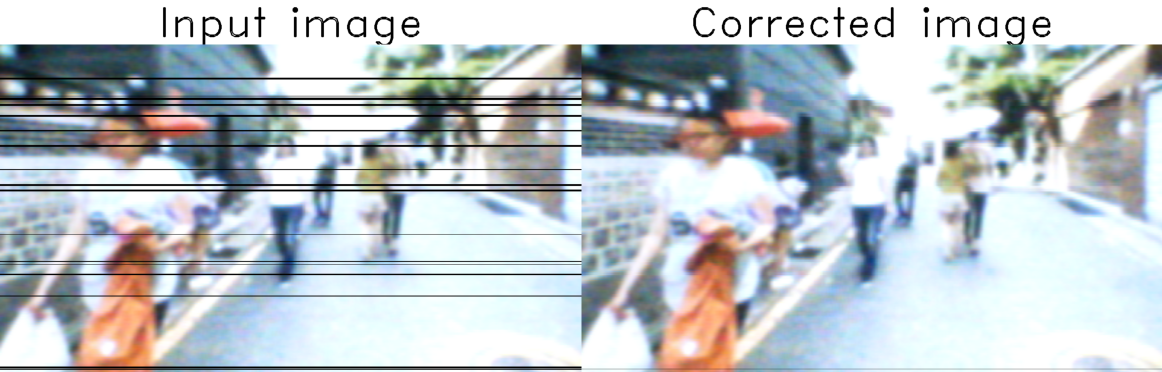}
    \end{subfigure}
    \begin{subfigure}[b]{0.9\linewidth}
        \centering
        \includegraphics[width=\linewidth]{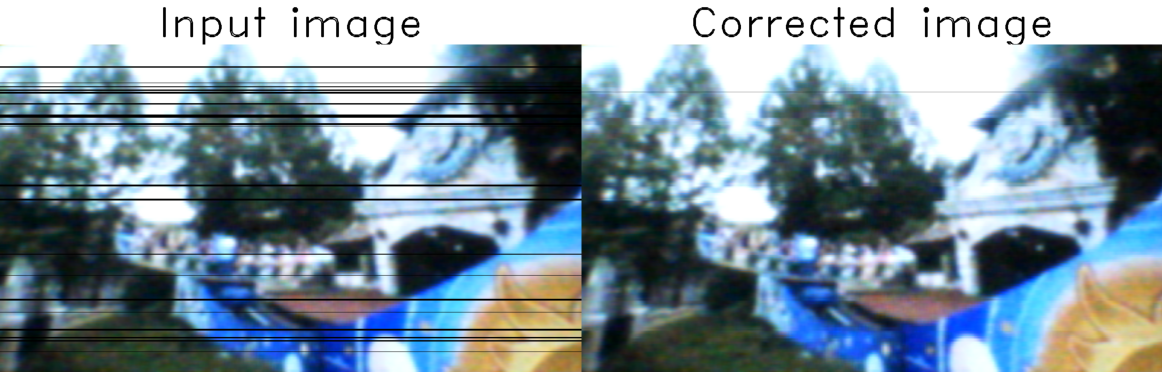}
    \end{subfigure}
    \vskip -0.12in
    \caption{Examples of our  packet loss correction.}
    \vskip -0.15in
    \label{fig:packet_loss}
\end{figure}


\vskip 0.05in\noindent{\bf Model run-time analysis.} 
At 15~fps, the receiver must process each frame in less than 66~ms. We measure the model runtime 
on a Nvidia RTX 2080 Ti GPU where it took less than 54~ms per frame. Since our model operates on key-frame to key-frame sequences, the minimum latency of our system is equal
to the period of the key-frame, i.e., 1~s.  Running inference on
our model with multiple batches  at a time increases the GPU utilization and  could improve the overall throughput
of our model. That is, multiple key-frame to key-frame sequences can be stacked along the batch
dimension to improve the throughput of the model. We observe significant throughput gains until a batch size being 4, where the runtime per frame is around 44~ms. We observe
no noticeable  gains when increase the batch size to more than 4.


\vskip 0.05in\noindent{\bf Power analysis.}  Table.~\ref{table:analysispower} shows the average current for the various hardware components. Since we do not perform compression on our device prototype, the radios draw more  current to transmit  to the base station. In the next section, we explore the design space with on-device  compression.


\begin{table}[t!]
\caption{{Power consumption of different  components in camera hardware. *Camera requires 27.52mA, but  is only active for 40 ms and so is heavily duty-cycled.}}
\begin{center}
\vskip -0.15in
{\footnotesize
 \begin{tabular}{l c c}
 \toprule
 Components & Average current (mA)  & Voltage (V)\\ \hline
 \emph{Camera sensors} &&\\
 Himax HM01B0 (15fps)  & 0.97 & 2.8\\
 OmniVision OV7692 (1fps)& 1.10* & 2.8\\\hline
 \emph{On-device Computing} &&\\
 STM32L496 MCU & 4.57 & 1.8\\
 STM32U575 MCU & 18 & 1.8\\\hline
 \emph{Wireless} &&\\
 TI CC2640R2F Radios & 21.42 & 1.8\\ \hline
\end{tabular}
}
\label{table:analysispower}
\end{center}
\vskip -0.2in
\end{table}

\begin{table*}[t!]
\caption{{\bf Comparison of existing wireless video systems.} We  report active power when the system is  ON and transmitting. \cite{batteryvideo1} uses analog backscatter where the SNR of each pixel significantly reduces with distance due to its analog design. *Owing to our dual-mode design where the high resolution camera is heavily duty cycled, only 5.8~mW of this power is from our cameras (13x reduction), realizing our goal of making the camera sensor power negligible. The rest is consumed by the microcontrollers and radios, which we do not currently optimize in our system.}

\begin{center}
\vskip -0.15in
 \begin{tabular}{lccccc}
 \toprule
 & Active Power  & Resolution & Frame rate & Color & Comms\\ \hline
 BackCam~\cite{sachin1} & 9.7~mW & $160 \times  120$ & 1 fps & greyscale & backscatter\\
 BeetleCam~\cite{sciencerobotics_uw} &4.4-18 mW  & $160 \times  120$ & 1-5 fps & greyscale & radio \\
  WISPCam~\cite{wispcam} & 6~mW & $176 \times  144$ & 0.001 fps & greyscale & backscatter \\
  Video Backscatter~\cite{batteryvideo1} & \splitr{2.4~mW\\
 - (sim)} 
  & \splitr{$112\times 112$\\
  1080p (sim)}
  & \splitr{13 fps\\ 10-60 fps} & \splitr{greyscale\\color} & \splitr{analog \\ backscatter} \\
  Blink~\cite{blink_power} & 700-1800~mW & $640\times 480$/740p & 7.5-30 fps & color & radio  \\
  Our prototype & {85~mW}* & $640 \times 480$ & 15 fps & color & radio \\
 \hline
\end{tabular}
\label{table:systable}
\end{center}
\vskip -0.15in
\end{table*}


\section{Related Work}

\textbf{Energy-efficiency and cameras.}   There has been recent work on designing more power-efficient camera sensors~\cite{camerapower1,camerapower2,michigan}, mixed-signal vision integrated circuits~\cite{mixed1,mixed2} and  processors~\cite{processors1,processors2}. The general relation between power consumption versus resolution, color, and SNR however still applies. 
\cite{745120} uses a combination of image compression  and simple pixel-difference-based motion detection on the receiver to compress the input video. \cite{726968} on the other hand does not use image compression but implements a remote control mechanism that can trigger the transmission of a single image.  \cite{1693344, 4415154} show medical   wireless cameras.  

Many commercial wireless cameras use on-device compression to reduce the data rate, including  recent work that uses neural networks for on-device video compression~\cite{wu2018vcii,Rippel2019LearnedVC,8953892,Shirkoohi2021EfficientVC}. In contrast to internet video streaming, low-power IoT  cameras on the other hand have reversed complexity, where the encoder is extremely resource-constrained, but the decoder can be relatively complex. While low-power accelerators have been proposed for compression, they are limited to images~\cite{starfish,michigan}. \cite{6142537, 7422694,glimpse} use low-power image compression, and also serve as good  references for limits on the power consumption of compression-based techniques. We focus on the camera sensor power consumption  and  explore the use of key-frame-based super-resolution and colorization.  Compression is complementary to sensor power and can also be applied to our low-power dual camera system.

 \cite{sciencerobotics_uw} presents a Bluetooth-based video streaming system. \cite{sachin1,batteryvideo1,batteryvideo2}  proposed backscatter systems for achieving low-power wireless video (see Table.~\ref{table:systable}). Backscatter is complementary to our approach and can be used instead of radios. 

\noindent\textbf{Neural networks for video processing.}  Super-resolution   has gained renewed popularity with the effectiveness of deep learning methods \cite{10.5555/2969239.2969266, 8099502, 8578791, 8237536, 7780576, 8099787, Ignatov2021RealTimeVS}. Variations to traditional convolution operations like sub-pixel convolutions \cite{7780576}, deformable convolutions \cite{8237351, 8953797, tian2020tdan, 8237351, wang2019edvr,8578438,10.5555/3157096.3157171} have been shown to work well for video super-resolution. However, more recently, it has been shown that recurrent networks \cite{RBPN2019, 10.1007/978-3-030-58610-2_38, chan2021basicvsr} can outperform  feed-forward networks that operate on a small batch of frames, at much lower complexity. \cite{chan2021basicvsr}  performs  super-resolution on these low-resolution frames and the rest of the frames are super-resolved using motion information and residual present in the encoded bit stream referring to super-resolved key-frames.  These works perform super-resolution using only low-resolution frames and hence cannot extract high-frequency details that may be non-existent in the low-resolution input. {\cite{10.1145/3152434.3152440, 10.1145/3372224.3419185, 10.1145/3387514.3405856} explore deep learning  for internet video streaming  to consumer devices. In contrast,  we work on the reverse  problem of continuous video capture from low-power IoT  cameras.}


While reference-based {\it image} super-resolution has been explored~\cite{Zheng2017LearningCC, zheng2018crossnet},  key-frame based {\it video}  super-resolution has not received the same attention. Early work on this considers this problem as a special case of the example-based super-resolution problem  \cite{988747}, where a low-resolution image or video is super-resolved based on a pool of high-resolution images. While generalized block matching methods  have been proposed for the example-based super-resolution problem \cite{988747, 6489325}, works on key-frame-based super-resolution specialize those methods by performing motion-compensated block matching. Those methods, however, only achieve a 2x scaling  and do not consider color~\cite{6205616, 4711756, 5604296}. Recent related work  on reference-based super-resolution (RefSR) \cite{Zheng2017LearningCC, zheng2018crossnet, 9328132, 9157498, inproceedings} only consider images,  not video, and do not transfer color.  Our work presents the first deep-learning technique to tackle  key-frame-based video super-resolution and colorization.

 Early colorization  techniques  propagated user
 suggestions to other parts of an image using optimization  
\cite{10.1145/1015706.1015780, 10.5555/2383847.2383887}. More recently, deep learning
methods have been  more effective ~\cite{7410412, Endo2016DeepPropED, 10.1145/2897824.2925974,
8100206,10.1007/978-3-030-58610-2_38, zhang2017real}. These methods are primarily
designed for image editing applications and need color hints at precise locations.
As a result, they tend to lose temporal coherence within a few frames \cite{Meyer18Color}.
 Welsh et al. \cite{1693344},
suggested that color spaces that separate luminance and color channels work better than
RGB color space. In recent years,  deep learning methods have been proposed for
color propagation  to achieve reliable transfer of color for a large number of
frames \cite{jampani:vpn:2017, 10.1007/978-3-319-46487-9_39, Meyer18Color}. These methods
operate on input videos in original resolution and as a result, do not perform scaling
on the luminance channel. Further,  we show that our two-way propagation mechanism achieves a much higher PSNR.

Prior work also designed neural networks for dual-camera settings~\cite{Cheng2021ADC}.   \cite{stereo} improves  stereo image super-resolution by using the symmetry cues in stereo image pairs. 
\cite{Wang2021DualCameraSW} targets super-resolution for  the wide-angle and telephoto cameras on phones that capture a scene with different focal lengths. {This method  only operates on single images at a time and is conceptually similar to  RefSR.}
 \cite{new} proposes a three-color camera system for super-resolving a low-resolution ultra-wide video utilizing wide-angle and telephoto videos. Our formulation differs in that  \cite{new} assume all three video streams are running at the same frame rate, whereas in our work, low-resolution and high-resolution cameras operate at significantly different frame rates. Moreover, they do not consider simultaneous super-resolution and colorization.

\vskip 0.25in\section{Conclusion and discussion}
\label{conclusion}
We introduce a novel deep learning-based system that can capture video from low-power dual-mode  IoT camera  systems. Here, we discuss some of our design tradeoffs.

\vskip 0.03in\noindent{\bf Power asymmetry.} We reduce the power consumption at the camera device while using a neural network decoder at the plugged-in router. The IoT devices are typically battery-powered and hence are more power-constrained than the plugged-in routers. We exploit this asymmetry to reduce power consumption at the IoT device.  We believe that this design choice is reasonable given that 1) the plugged-in router is not power-constrained like the IoT camera, and 2) neural accelerator chips are being developed to run  neural networks more energy-efficiently than using  GPUs. 

\vskip 0.03in\noindent{\bf Design space.} There are a few other design points that can leverage this power asymmetry. However, we found in our experiments that they provide lower-quality output while being practically harder to implement. Sensing only low-resolution color video and super-resolving it on the receiver is one such approach. As we observed in Fig. \ref{fig:3c_comp_y}, this results in a significant drop in performance. Furthermore, commercial color cameras have a large active pixel area supporting multiple resolutions. The low-resolution color video would then have to be obtained by artificially sub-sampling during the read-out, thus not realizing the similar power savings that purpose-built low-resolution grayscale sensors provide. Another approach is to use  super-resolution followed by color propagation on the receiver, using existing computer vision models for those tasks. In this case, only color channels are required to be transmitted. This can partially save the communication power but requires color space conversion on the device. Moreover, discarding the high-resolution luminance channel would be a sub-optimal design  given the   gains that high-resolution details in the key-frames provide.

\vskip 0.03in\noindent{\bf Latency tradeoff.} In our bi-directional network, we use the next key-frame to be able to perform super-resolution, which adds a fixed latency. One approach is to use a playback buffer, similar to video on demand like Youtube,  to address this latency. We could also add a uni-directional path to the network for low-latency uses and improve the resolution with our bi-directional network for on-demand  retrieval of the video data as is common in security camera applications. We note that in sensor applications like wildlife monitoring and smart farms, a key-frame latency is acceptable.

\vskip 0.03in\noindent {\bf Applications and limitations.} Our work has applications in both live video blogging (wearable cameras, GoPro, etc.) and security monitoring. Compared to traditional video super-resolution methods  applied for these applications, our method shows improved robustness by using accurate high-frequency details present in key-frames. In addition, key-frames also can be preserved and referred to in case of ambiguity. This does not eliminate the possibility of objects appearing and disappearing before a key-frame capture them, which is a valid limitation of our current approach. However, future works based on this method could include intelligent algorithms, e.g., based on reinforcement learning, to detect the optimal moment to capture a key-frame.

\vskip 0.03in\noindent{\bf Societal considerations.} Since the  video is produced by a deep learning model, it is important to ponder on  the trust a user may be willing to place in the output, e.g., as legal evidence. These models may  also  be open to adversarial attacks. In our system, since the key-frames which have color and high resolution are stored, they can be referred to in case of ambiguity. To make it  more robust, a possible future work is to adaptively capture more key-frames after learning when an event of interest occurs.

\vskip 0.05in\noindent{\bf Acknowledgments.}  The  researchers are funded by the Moore Inventor Fellow award \#10617 and NSF.

\bibliographystyle{plain}
\bibliography{refs}

\begin{thebibliography}{10}

\bibitem{hicamera-spec}
High-speed vga image sensor for video, gaming and portable entertainment.
\newblock \url{
  https://pdf1.alldatasheet.net/datasheet-pdf/view/981148/OMNIVISION/OV7735.html}.

\bibitem{locamera-spec}
Hm01b0-mna-01ft870 compact camera module.
\newblock \url{
  https://cdn.sparkfun.com/assets/7/f/c/8/3/HM01B0-MNA-Datasheet.pdf}.

\bibitem{blink_power}
Investigating power consumption of blink.
\newblock
  \url{https://cam-do.com/blogs/camdo-blog/investigating-power-consumption-of-blink}.

\bibitem{h264_ic}
Mb86h45/mb86h46 ultra low power full hd h.264 codecs.
\newblock
  \url{https://www.fujitsu.com/downloads/MICRO/fme/documentation/m24.pdf}.

\bibitem{cmosnoise}
Control of noise and background in scientific cmos technology, teledyne
  imaging, 2019.
\newblock
  \url{https://www.photometrics.com/wp-content/uploads/2019/11/Control-of-Noise-and-Background-in-Scientific-CMOS-Technology.pdf}.

\bibitem{picamera2}
{Raspberry Pi Camera Module 2},.
\newblock \url{https://www.raspberrypi.com/products/camera-module-v2/}, 2022.

\bibitem{726968}
M.J. Agan, B.H. Olson, C.R. Pasqualino, and G.L. Stevens.
\newblock A highly miniaturized, battery operated, commandable, digital
  wireless camera.
\newblock In {\em IEEE Military Communications Conference. Proceedings. MILCOM
  98 (Cat. No.98CH36201)}, volume~3, pages 913--918 vol.3, 1998.

\bibitem{4914842}
Asral Bahari, Tughrul Arslan, and Ahmet~T. Erdogan.
\newblock Low-power h.264 video compression architectures for mobile
  communication.
\newblock {\em IEEE Transactions on Circuits and Systems for Video Technology},
  19(9):1251--1261, 2009.

\bibitem{Bahdanau2015NeuralMT}
Dzmitry Bahdanau, Kyunghyun Cho, and Yoshua Bengio.
\newblock Neural machine translation by jointly learning to align and
  translate.
\newblock {\em CoRR}, abs/1409.0473, 2015.

\bibitem{michigan}
Andrea Bejarano-Carbo, Hyochan An, Kyojin Choo, Shiyu Liu, Dennis Sylvester,
  David Blaauw, and Hun-Seok Kim.
\newblock Millimeter-scale ultra-low-power imaging system for intelligent edge
  monitoring, 2022.

\bibitem{4711756}
Fernanda Brandi, Ricardo de~Queiroz, and Debargha Mukherjee.
\newblock Super-resolution of video using key frames and motion estimation.
\newblock In {\em 2008 15th IEEE International Conference on Image Processing},
  pages 321--324, 2008.

\bibitem{8099787}
Jose Caballero, Christian Ledig, Andrew Aitken, Alejandro Acosta, Johannes
  Totz, Zehan Wang, and Wenzhe Shi.
\newblock Real-time video super-resolution with spatio-temporal networks and
  motion compensation.
\newblock In {\em 2017 IEEE Conference on Computer Vision and Pattern
  Recognition (CVPR)}, pages 2848--2857, 2017.

\bibitem{chan2021basicvsr}
Kelvin~CK Chan, Xintao Wang, Ke~Yu, Chao Dong, and Chen~Change Loy.
\newblock Basicvsr: The search for essential components in video
  super-resolution and beyond.
\newblock In {\em Proceedings of the IEEE conference on computer vision and
  pattern recognition}, 2021.

\bibitem{chan2022basicvsrplusplus}
Kelvin~C.K. Chan, Shangchen Zhou, Xiangyu Xu, and Chen~Change Loy.
\newblock Basicvsr++: Improving video super-resolution with enhanced
  propagation and alignment.
\newblock In {\em Proceedings of the IEEE conference on computer vision and
  pattern recognition}, 2022.

\bibitem{chandra2021smart}
Ranveer Chandra and Stewart Collis.
\newblock Smart farming with technologies such as iot, computer vision, and ai
  can improve agricultural efficiency, transparency, profitability, and equity
  for farmers in low-and middle-income countries.
\newblock {\em COMMUNICATIONS OF THE ACM}, 64(12):75--84, 2021.

\bibitem{745120}
A.~Chandrakasan, A.~Dancy, J.~Goodman, and T.~Simon.
\newblock A low-power wireless camera system.
\newblock In {\em Proceedings Twelfth International Conference on VLSI Design.
  (Cat. No.PR00013)}, pages 32--36, 1999.

\bibitem{413553}
P.~Charbonnier, L.~Blanc-Feraud, G.~Aubert, and M.~Barlaud.
\newblock Two deterministic half-quadratic regularization algorithms for
  computed imaging.
\newblock In {\em Proceedings of 1st International Conference on Image
  Processing}, volume~2, pages 168--172 vol.2, 1994.

\bibitem{Cheng2021ADC}
Ming Cheng, Zhan Ma, M.~Salman Asif, Yiling Xu, Haojie Liu, Wenbo Bao, and Jun
  Sun.
\newblock A dual camera system for high spatiotemporal resolution video
  acquisition.
\newblock {\em IEEE Transactions on Pattern Analysis and Machine Intelligence},
  43:3275--3291, 2021.

\bibitem{7410412}
Zezhou Cheng, Qingxiong Yang, and Bin Sheng.
\newblock Deep colorization.
\newblock In {\em 2015 IEEE International Conference on Computer Vision
  (ICCV)}, pages 415--423, 2015.

\bibitem{8237351}
Jifeng Dai, Haozhi Qi, Yuwen Xiong, Yi~Li, Guodong Zhang, Han Hu, and Yichen
  Wei.
\newblock Deformable convolutional networks.
\newblock In {\em 2017 IEEE International Conference on Computer Vision
  (ICCV)}, pages 764--773, 2017.

\bibitem{10.5555/3157096.3157171}
Bert De~Brabandere, Xu~Jia, Tinne Tuytelaars, and Luc Van~Gool.
\newblock Dynamic filter networks.
\newblock In {\em Proceedings of the 30th International Conference on Neural
  Information Processing Systems}, NIPS'16, page 667–675, Red Hook, NY, USA,
  2016. Curran Associates Inc.

\bibitem{9328132}
Runmin Dong, Lixian Zhang, and Haohuan Fu.
\newblock Rrsgan: Reference-based super-resolution for remote sensing image.
\newblock {\em IEEE Transactions on Geoscience and Remote Sensing}, pages
  1--17, 2021.

\bibitem{dosovitskiy2020vit}
Alexey Dosovitskiy, Lucas Beyer, Alexander Kolesnikov, Dirk Weissenborn,
  Xiaohua Zhai, Thomas Unterthiner, Mostafa Dehghani, Matthias Minderer, Georg
  Heigold, Sylvain Gelly, Jakob Uszkoreit, and Neil Houlsby.
\newblock An image is worth 16x16 words: Transformers for image recognition at
  scale.
\newblock {\em ICLR}, 2021.

\bibitem{Endo2016DeepPropED}
Y.~Endo, S.~Iizuka, Y.~Kanamori, and J.~Mitani.
\newblock Deepprop: Extracting deep features from a single image for edit
  propagation.
\newblock {\em Computer Graphics Forum}, 35, 2016.

\bibitem{988747}
W.T. Freeman, T.R. Jones, and E.C. Pasztor.
\newblock Example-based super-resolution.
\newblock {\em IEEE Computer Graphics and Applications}, 22(2):56--65, 2002.

\bibitem{RBPN2019}
Muhammad Haris, Greg Shakhnarovich, and Norimichi Ukita.
\newblock Recurrent back-projection network for video super-resolution.
\newblock In {\em IEEE Conference on Computer Vision and Pattern Recognition
  (CVPR)}, 2019.

\bibitem{10.1145/3197517.3201365}
Mingming He, Dongdong Chen, Jing Liao, Pedro~V. Sander, and Lu~Yuan.
\newblock Deep exemplar-based colorization.
\newblock {\em ACM Trans. Graph.}, 37(4), jul 2018.

\bibitem{starfish}
Pan Hu, Junha Im, Zain Asgar, and Sachin Katti.
\newblock {\em Starfish: Resilient Image Compression for AIoT Cameras}, page
  395–408.
\newblock Association for Computing Machinery, New York, NY, USA, 2020.

\bibitem{10.5555/2969239.2969266}
Yan Huang, Wei Wang, and Liang Wang.
\newblock Bidirectional recurrent convolutional networks for multi-frame
  super-resolution.
\newblock In {\em Proceedings of the 28th International Conference on Neural
  Information Processing Systems - Volume 1}, NIPS'15, page 235–243,
  Cambridge, MA, USA, 2015. MIT Press.

\bibitem{6205616}
Edson~Mintsu Hung, Ricardo~L. de~Queiroz, Fernanda Brandi, Karen~França
  de~Oliveira, and Debargha Mukherjee.
\newblock Video super-resolution using codebooks derived from key-frames.
\newblock {\em IEEE Transactions on Circuits and Systems for Video Technology},
  22(9):1321--1331, 2012.

\bibitem{Ignatov2021RealTimeVS}
Andrey~D. Ignatov, Andr{\'e}s Romero, Heewon Kim, Radu Timofte, Chiu~Man Ho,
  Zibo Meng, Kyoung~Mu Lee, Yuxiang Chen, Yutong Wang, Zeyu Long, Chenhao Wang,
  Yifei Chen, Boshen Xu, Shuhang Gu, Lixin Duan, Wen Li, Wang Bofei, Zhang
  Diankai, Zheng Chengjian, Liu Shaoli, Gao Si, Zhang Xiaofeng, Luan Kaidi,
  Xu~Tianyu, Zheng Hui, Xinbo Gao, Xiumei Wang, Jiaming Guo, Xueyi Zhou, Hao
  Jia, and Youliang Yan.
\newblock Real-time video super-resolution on smartphones with deep learning,
  mobile ai 2021 challenge: Report.
\newblock {\em 2021 IEEE/CVF Conference on Computer Vision and Pattern
  Recognition Workshops (CVPRW)}, pages 2535--2544, 2021.

\bibitem{10.1145/2897824.2925974}
Satoshi Iizuka, Edgar Simo-Serra, and Hiroshi Ishikawa.
\newblock Let there be color! joint end-to-end learning of global and local
  image priors for automatic image colorization with simultaneous
  classification.
\newblock {\em ACM Trans. Graph.}, 35(4), July 2016.

\bibitem{10.1007/978-3-030-58610-2_38}
Takashi Isobe, Xu~Jia, Shuhang Gu, Songjiang Li, Shengjin Wang, and Qi~Tian.
\newblock Video super-resolution with recurrent structure-detail network.
\newblock In Andrea Vedaldi, Horst Bischof, Thomas Brox, and Jan-Michael Frahm,
  editors, {\em Computer Vision -- ECCV 2020}, pages 645--660, Cham, 2020.
  Springer International Publishing.

\bibitem{1693344}
S.~Itoh, S.~Kawahito, and S.~Terakawa.
\newblock A 2.6mw 2fps qvga cmos one-chip wireless camera with digital image
  transmission function for capsule endoscopes.
\newblock In {\em 2006 IEEE International Symposium on Circuits and Systems},
  pages 4 pp.--3356, 2006.

\bibitem{sciencerobotics_uw}
Vikram Iyer, Ali Najafi, Johannes James, Sawyer Fuller, and Shyamnath
  Gollakota.
\newblock Wireless steerable vision for live insects and insect-scale robots.
\newblock {\em Science Robotics}, 5:eabb0839, 07 2020.

\bibitem{jampani:vpn:2017}
Varun Jampani, Raghudeep Gadde, and Peter~V. Gehler.
\newblock Video propagation networks.
\newblock In {\em IEEE Conf. on Computer Vision and Pattern Recognition
  (CVPR)}, july 2017.

\bibitem{camerapower1}
Suyao Ji, Jing Pu, Byong~Chan Lim, and Mark Horowitz.
\newblock A 220pj/pixel/frame cmos image sensor with partial settling readout
  architecture.
\newblock In {\em 2016 IEEE Symposium on VLSI Circuits (VLSI-Circuits)}, pages
  1--2, 2016.

\bibitem{8578438}
Younghyun Jo, Seoung~Wug Oh, Jaeyeon Kang, and Seon~Joo Kim.
\newblock Deep video super-resolution network using dynamic upsampling filters
  without explicit motion compensation.
\newblock In {\em 2018 IEEE/CVF Conference on Computer Vision and Pattern
  Recognition}, pages 3224--3232, 2018.

\bibitem{sachin1}
Colleen Josephson, Lei Yang, Pengyu Zhang, and Sachin Katti.
\newblock Wireless computer vision using commodity radios.
\newblock In {\em 2019 18th ACM/IEEE International Conference on Information
  Processing in Sensor Networks (IPSN)}, pages 229--240, 2019.

\bibitem{6142537}
Med~Lassaad Kaddachi, Leila Makkaoui, Adel Soudani, Vincent Lecuire, and
  Jean-Marie Moureaux.
\newblock Fpga-based image compression for low-power wireless camera sensor
  networks.
\newblock In {\em 2011 3rd International Conference on Next Generation Networks
  and Services (NGNS)}, pages 68--71, 2011.

\bibitem{10.1145/3387514.3405856}
Jaehong Kim, Youngmok Jung, Hyunho Yeo, Juncheol Ye, and Dongsu Han.
\newblock Neural-enhanced live streaming: Improving live video ingest via
  online learning.
\newblock SIGCOMM '20, page 107–125, New York, NY, USA, 2020. Association for
  Computing Machinery.

\bibitem{article}
Diederik Kingma and Jimmy Ba.
\newblock Adam: A method for stochastic optimization.
\newblock {\em International Conference on Learning Representations}, 12 2014.

\bibitem{wilds}
Pang~Wei Koh, Shiori Sagawa, Henrik Marklund, Sang~Michael Xie, Marvin Zhang,
  Akshay Balsubramani, Weihua Hu, Michihiro Yasunaga, Richard~Lanas Phillips,
  Irena Gao, Tony Lee, Etienne David, Ian Stavness, Wei Guo, Berton Earnshaw,
  Imran Haque, Sara~M Beery, Jure Leskovec, Anshul Kundaje, Emma Pierson,
  Sergey Levine, Chelsea Finn, and Percy Liang.
\newblock Wilds: A benchmark of in-the-wild distribution shifts.
\newblock In Marina Meila and Tong Zhang, editors, {\em Proceedings of the 38th
  International Conference on Machine Learning}, volume 139 of {\em Proceedings
  of Machine Learning Research}, pages 5637--5664. PMLR, 18--24 Jul 2021.

\bibitem{dinesh}
Shih-Kai Kuo, Manideep Dunna, Dinesh Bharadia, and Patrick~P. Mercier.
\newblock A wifi and bluetooth backscattering combo chip featuring beam
  steering via a fully-reflective phased-controlled multi-antenna termination
  technique enabling operation over 56 meters.
\newblock In {\em 2022 IEEE International Solid- State Circuits Conference
  (ISSCC)}, volume~65, pages 1--3, 2022.

\bibitem{8099502}
Christian Ledig, Lucas Theis, Ferenc Huszár, Jose Caballero, Andrew
  Cunningham, Alejandro Acosta, Andrew Aitken, Alykhan Tejani, Johannes Totz,
  Zehan Wang, and Wenzhe Shi.
\newblock Photo-realistic single image super-resolution using a generative
  adversarial network.
\newblock In {\em 2017 IEEE Conference on Computer Vision and Pattern
  Recognition (CVPR)}, pages 105--114, 2017.

\bibitem{new}
Junyong Lee, Myeonghee Lee, Sunghyun Cho, and Seungyong Lee.
\newblock Reference-based video super-resolution using multi-camera video
  triplets, March 2022, arxiv.

\bibitem{mixed1}
Martin Lefebvre, Ludovic Moreau, Rémi Dekimpe, and David Bol.
\newblock 7.7 a 0.2-to-3.6tops/w programmable convolutional imager soc with
  in-sensor current-domain ternary-weighted mac operations for feature
  extraction and region-of-interest detection.
\newblock In {\em 2021 IEEE International Solid- State Circuits Conference
  (ISSCC)}, volume~64, pages 118--120, 2021.

\bibitem{10.1145/1015706.1015780}
Anat Levin, Dani Lischinski, and Yair Weiss.
\newblock Colorization using optimization.
\newblock {\em ACM Trans. Graph.}, 23(3):689–694, August 2004.

\bibitem{powerscaling}
Robert LiKamWa, Bodhi Priyantha, Matthai Philipose, Lin Zhong, and Paramvir
  Bahl.
\newblock Energy characterization and optimization of image sensing toward
  continuous mobile vision.
\newblock In {\em Proceeding of the 11th Annual International Conference on
  Mobile Systems, Applications, and Services}, MobiSys '13, page 69–82, New
  York, NY, USA, 2013. Association for Computing Machinery.

\bibitem{6549107}
Ce~Liu and Deqing Sun.
\newblock On bayesian adaptive video super resolution.
\newblock {\em IEEE Transactions on Pattern Analysis and Machine Intelligence},
  36(2):346--360, 2014.

\bibitem{8237536}
Ding Liu, Zhaowen Wang, Yuchen Fan, Xianming Liu, Zhangyang Wang, Shiyu Chang,
  and Thomas Huang.
\newblock Robust video super-resolution with learned temporal dynamics.
\newblock In {\em 2017 IEEE International Conference on Computer Vision
  (ICCV)}, pages 2526--2534, 2017.

\bibitem{Liu_cmosimage}
Xinqiao Liu, Abbas~El Gamal, Mark~A. Horowitz, and Brian~A. Wandell.
\newblock Cmos image sensors dynamic range and snr enhancement via statistical
  signal processing, stanford ph.d. thesis, 2002.

\bibitem{8953892}
Guo Lu, Wanli Ouyang, Dong Xu, Xiaoyun Zhang, Chunlei Cai, and Zhiyong Gao.
\newblock Dvc: An end-to-end deep video compression framework.
\newblock In {\em 2019 IEEE/CVF Conference on Computer Vision and Pattern
  Recognition (CVPR)}, pages 10998--11007, 2019.

\bibitem{10.5555/2383847.2383887}
Qing Luan, Fang Wen, Daniel Cohen-Or, Lin Liang, Ying-Qing Xu, and Heung-Yeung
  Shum.
\newblock Natural image colorization.
\newblock In {\em Proceedings of the 18th Eurographics Conference on Rendering
  Techniques}, EGSR'07, page 309–320, Goslar, DEU, 2007. Eurographics
  Association.

\bibitem{DBLP:journals/corr/abs-2110-02178}
Sachin Mehta and Mohammad Rastegari.
\newblock Mobilevit: Light-weight, general-purpose, and mobile-friendly vision
  transformer.
\newblock {\em CoRR}, abs/2110.02178, 2021.

\bibitem{Meyer18Color}
Simone Meyer, Victor Cornill\`ere, Abdelaziz Djelouah, Christopher Schroers,
  and Markus Gross.
\newblock Deep video color propagation.
\newblock In {\em Proceedings of the British Machine Vision Conference {BMVC}},
  2018.

\bibitem{10.1007/978-3-319-46487-9_39}
Simone Meyer, Alexander Sorkine-Hornung, and Markus Gross.
\newblock Phase-based modification transfer for video.
\newblock In Bastian Leibe, Jiri Matas, Nicu Sebe, and Max Welling, editors,
  {\em Computer Vision -- ECCV 2016}, pages 633--648, Cham, 2016. Springer
  International Publishing.

\bibitem{camerapower2}
Fukashi Morishita, Norihito Kato, Satoshi Okubo, Takao Toi, Mitsuru Hiraki,
  Sugako Otani, Hideaki Abe, Yuji Shinohara, and Hiroyuki Kondo.
\newblock A cmos image sensor and an ai accelerator for realizing
  edge-computing-based surveillance camera systems.
\newblock In {\em 2021 Symposium on VLSI Circuits}, pages 1--2, 2021.

\bibitem{batteryvideo1}
Saman Naderiparizi, Mehrdad Hessar, Vamsi Talla, Shyamnath Gollakota, and
  Joshua~R. Smith.
\newblock Towards battery-free hd video streaming.
\newblock In {\em Proceedings of the 15th USENIX Conference on Networked
  Systems Design and Implementation}, NSDI'18, page 233–247, USA, 2018.
  USENIX Association.

\bibitem{wispcam}
Saman Naderiparizi, Aaron~N. Parks, Zerina Kapetanovic, Benjamin Ransford, and
  Joshua~R. Smith.
\newblock Wispcam: A battery-free rfid camera.
\newblock In {\em 2015 IEEE International Conference on RFID (RFID)}, pages
  166--173, 2015.

\bibitem{glimpse}
Saman Naderiparizi, Pengyu Zhang, Matthai Philipose, Bodhi Priyantha, Jie Liu,
  and Deepak Ganesan.
\newblock Glimpse: A programmable early-discard camera architecture for
  continuous mobile vision.
\newblock In {\em Proceedings of the 15th Annual International Conference on
  Mobile Systems, Applications, and Services}, MobiSys '17, page 292–305, New
  York, NY, USA, 2017. Association for Computing Machinery.

\bibitem{Nah_2019_CVPR_Workshops_REDS}
Seungjun Nah, Sungyong Baik, Seokil Hong, Gyeongsik Moon, Sanghyun Son, Radu
  Timofte, and Kyoung~Mu Lee.
\newblock Ntire 2019 challenge on video deblurring and super-resolution:
  Dataset and study.
\newblock In {\em CVPR Workshops}, June 2019.

\bibitem{6489325}
Seyedreza Najafi and Shahram Shirani.
\newblock Regularization function for video super-resolution using auxiliary
  high resolution still images.
\newblock In {\em 2012 Conference Record of the Forty Sixth Asilomar Conference
  on Signals, Systems and Computers (ASILOMAR)}, pages 1713--1717, 2012.

\bibitem{videoio}
Nuo Nan, Haian Zhou, Xin Li, Qi~Tang, Xingtai Feng, and Yuan Yuan.
\newblock {Design of a light and high-resolution video camera system based on
  CMV12000 sensor}.
\newblock In Junhong Su, Lianghui Chen, Junhao Chu, Shining Zhu, and Qifeng Yu,
  editors, {\em Eighth Symposium on Novel Photoelectronic Detection Technology
  and Applications}, volume 12169, pages 2619 -- 2625. International Society
  for Optics and Photonics, SPIE, 2022.

\bibitem{4415154}
Lorenzo Piccardi, Basilio Noris, Olivier Barbey, Aude Billard, Giuseppina
  Schiavone, Flavio Keller, and Claes von Hofsten.
\newblock Wearcam: A head mounted wireless camera for monitoring gaze attention
  and for the diagnosis of developmental disorders in young children.
\newblock In {\em RO-MAN 2007 - The 16th IEEE International Symposium on Robot
  and Human Interactive Communication}, pages 594--598, 2007.

\bibitem{Rippel2019LearnedVC}
Oren Rippel, Sanjay Nair, Carissa Lew, Steve Branson, Alexander~G. Anderson,
  and Lubomir~D. Bourdev.
\newblock Learned video compression.
\newblock {\em 2019 IEEE/CVF International Conference on Computer Vision
  (ICCV)}, pages 3453--3462, 2019.

\bibitem{processors1}
Davide Rossi, Francesco Conti, Manuel Eggiman, Stefan Mach, Alfio~Di Mauro,
  Marco Guermandi, Giuseppe Tagliavini, Antonio Pullini, Igor Loi, Jie Chen,
  Eric Flamand, and Luca Benini.
\newblock 4.4 a 1.3tops/w @ 32gops fully integrated 10-core soc for iot
  end-nodes with 1.7uw cognitive wake-up from mram-based state-retentive sleep
  mode.
\newblock In {\em 2021 IEEE International Solid- State Circuits Conference
  (ISSCC)}, volume~64, pages 60--62, 2021.

\bibitem{batteryvideo2}
Ali Saffari, Mehrdad Hessar, Saman Naderiparizi, and Joshua~R. Smith.
\newblock Battery-free wireless video streaming camera system.
\newblock In {\em 2019 IEEE International Conference on RFID (RFID)}, pages
  1--8, 2019.

\bibitem{8578791}
Mehdi S.~M. Sajjadi, Raviteja Vemulapalli, and Matthew Brown.
\newblock Frame-recurrent video super-resolution.
\newblock In {\em 2018 IEEE/CVF Conference on Computer Vision and Pattern
  Recognition}, pages 6626--6634, 2018.

\bibitem{8100206}
Patsorn Sangkloy, Jingwan Lu, Chen Fang, Fisher Yu, and James Hays.
\newblock Scribbler: Controlling deep image synthesis with sketch and color.
\newblock In {\em 2017 IEEE Conference on Computer Vision and Pattern
  Recognition (CVPR)}, pages 6836--6845, 2017.

\bibitem{10.1145/3115505}
Yousef~O. Sharrab and Nabil~J. Sarhan.
\newblock Modeling and analysis of power consumption in live video streaming
  systems.
\newblock {\em ACM Trans. Multimedia Comput. Commun. Appl.}, 13(4), sep 2017.

\bibitem{7780576}
Wenzhe Shi, Jose Caballero, Ferenc Huszár, Johannes Totz, Andrew~P. Aitken,
  Rob Bishop, Daniel Rueckert, and Zehan Wang.
\newblock Real-time single image and video super-resolution using an efficient
  sub-pixel convolutional neural network.
\newblock In {\em 2016 IEEE Conference on Computer Vision and Pattern
  Recognition (CVPR)}, pages 1874--1883, 2016.

\bibitem{9157498}
Gyumin Shim, Jinsun Park, and In~So Kweon.
\newblock Robust reference-based super-resolution with similarity-aware
  deformable convolution.
\newblock In {\em 2020 IEEE/CVF Conference on Computer Vision and Pattern
  Recognition (CVPR)}, pages 8422--8431, 2020.

\bibitem{Shirkoohi2021EfficientVC}
Mehrdad~Khani Shirkoohi, Vibhaalakshmi Sivaraman, and Mohammad Alizadeh.
\newblock Efficient video compression via content-adaptive super-resolution.
\newblock {\em 2021 IEEE/CVF International Conference on Computer Vision
  (ICCV)}, pages 4501--4510, 2021.

\bibitem{processors2}
Sander Smets, Toon Goedemé, Anurag Mittal, and Marian Verhelst.
\newblock 2.2 a 978gops/w flexible streaming processor for real-time image
  processing applications in 22nm fdsoi.
\newblock In {\em 2019 IEEE International Solid- State Circuits Conference -
  (ISSCC)}, pages 44--46, 2019.

\bibitem{5604296}
Byung~Cheol Song, Shin-Cheol Jeong, and Yanglim Choi.
\newblock Video super-resolution algorithm using bi-directional overlapped
  block motion compensation and on-the-fly dictionary training.
\newblock {\em IEEE Transactions on Circuits and Systems for Video Technology},
  21(3):274--285, 2011.

\bibitem{10.1145/3130970}
Vamsi Talla, Mehrdad Hessar, Bryce Kellogg, Ali Najafi, Joshua~R. Smith, and
  Shyamnath Gollakota.
\newblock Lora backscatter: Enabling the vision of ubiquitous connectivity.
\newblock {\em Proc. ACM Interact. Mob. Wearable Ubiquitous Technol.}, 1(3),
  sep 2017.

\bibitem{powifi}
Vamsi Talla, Bryce Kellogg, Benjamin Ransford, Saman Naderiparizi, Shyamnath
  Gollakota, and Joshua Smith.
\newblock Powering the next billion devices with wi-fi.
\newblock In {\em ACM CONEXT}, pages 1--13, 12 2015.

\bibitem{tian2020tdan}
Yapeng Tian, Yulun Zhang, Yun Fu, and Chenliang Xu.
\newblock Tdan: Temporally-deformable alignment network for video
  super-resolution.
\newblock In {\em The IEEE Conference on Computer Vision and Pattern
  Recognition (CVPR)}, June 2020.

\bibitem{Wang2021DualCameraSW}
Tengfei Wang, Jiaxin Xie, Wenxiu Sun, Qiong Yan, and Qifeng Chen.
\newblock Dual-camera super-resolution with aligned attention modules.
\newblock {\em 2021 IEEE/CVF International Conference on Computer Vision
  (ICCV)}, pages 1981--1990, 2021.

\bibitem{wang2019edvr}
Xintao Wang, Kelvin~C.K. Chan, Ke~Yu, Chao Dong, and Chen~Change Loy.
\newblock Edvr: Video restoration with enhanced deformable convolutional
  networks.
\newblock In {\em The IEEE Conference on Computer Vision and Pattern
  Recognition (CVPR) Workshops}, June 2019.

\bibitem{stereo}
Yingqian Wang, Xinyi Ying, Longguang Wang, Jungang Yang, Wei An, and Yulan Guo.
\newblock Symmetric parallax attention for stereo image super-resolution.
\newblock {\em CoRR}, abs/2011.03802, 2020.

\bibitem{7422694}
Yong Wang, Dianhong Wang, Xufan Zhang, Jun Chen, and Yamin Li.
\newblock Energy-efficient image compressive transmission for wireless camera
  networks.
\newblock {\em IEEE Sensors Journal}, 16(10):3875--3886, 2016.

\bibitem{wu2018vcii}
Chao-Yuan Wu, Nayan Singhal, and Philipp Kr{\"a}henb{\"u}hl.
\newblock Video compression through image interpolation.
\newblock In {\em ECCV}, 2018.

\bibitem{mixed2}
Han Xu, Ziru Li, Ningchao Lin, Qi~Wei, Fei Qiao, Xunzhao Yin, and Huazhong
  Yang.
\newblock Macsen: A processing-in-sensor architecture integrating mac
  operations into image sensor for ultra-low-power bnn-based intelligent visual
  perception.
\newblock {\em IEEE Transactions on Circuits and Systems II: Express Briefs},
  68(2):627--631, 2021.

\bibitem{xue2019video}
Tianfan Xue, Baian Chen, Jiajun Wu, Donglai Wei, and William~T Freeman.
\newblock Video enhancement with task-oriented flow.
\newblock {\em International Journal of Computer Vision (IJCV)},
  127(8):1106--1125, 2019.

\bibitem{10.1145/3372224.3419185}
Hyunho Yeo, Chan~Ju Chong, Youngmok Jung, Juncheol Ye, and Dongsu Han.
\newblock Nemo: Enabling neural-enhanced video streaming on commodity mobile
  devices.
\newblock In {\em Proceedings of the 26th Annual International Conference on
  Mobile Computing and Networking}, MobiCom '20, New York, NY, USA, 2020.
  Association for Computing Machinery.

\bibitem{10.1145/3152434.3152440}
Hyunho Yeo, Sunghyun Do, and Dongsu Han.
\newblock How will deep learning change internet video delivery?
\newblock In {\em Proceedings of the 16th ACM Workshop on Hot Topics in
  Networks}, HotNets-XVI, page 57–64, New York, NY, USA, 2017. Association
  for Computing Machinery.

\bibitem{PFNL}
Peng Yi, Zhongyuan Wang, Kui Jiang, Junjun Jiang, and Jiayi Ma.
\newblock Progressive fusion video super-resolution network via exploiting
  non-local spatio-temporal correlations.
\newblock In {\em IEEE International Conference on Computer Vision (ICCV)},
  pages 3106--3115, 2019.

\bibitem{ekhonet}
Pengyu Zhang, Pan Hu, Vijay Pasikanti, and Deepak Ganesan.
\newblock Ekhonet: High-speed ultra low-power backscatter for next generation
  sensors.
\newblock {\em GetMobile Mob. Comput. Commun.}, 19:14--17, 2015.

\bibitem{zhang2017real}
Richard Zhang, Jun-Yan Zhu, Phillip Isola, Xinyang Geng, Angela~S Lin, Tianhe
  Yu, and Alexei~A Efros.
\newblock Real-time user-guided image colorization with learned deep priors.
\newblock {\em ACM Transactions on Graphics (TOG)}, 9(4), 2017.

\bibitem{inproceedings}
Zhifei Zhang, Zhaowen Wang, Zhe Lin, and Hairong Qi.
\newblock Image super-resolution by neural texture transfer.
\newblock pages 7974--7983, 06 2019.

\bibitem{Zheng2017LearningCC}
Haitian Zheng, Mengqi Ji, L.~Han, Z.~Xu, Haoqian Wang, Yebin Liu, and Lu~Fang.
\newblock Learning cross-scale correspondence and patch-based synthesis for
  reference-based super-resolution.
\newblock In {\em BMVC}, 2017.

\bibitem{zheng2018crossnet}
Haitian Zheng, Mengqi Ji, Haoqian Wang, Yebin Liu, and Lu~Fang.
\newblock Crossnet: An end-to-end reference-based super resolution network
  using cross-scale warping.
\newblock In {\em Proceedings of the European Conference on Computer Vision
  (ECCV)}, pages 88--104, 2018.

\bibitem{8953797}
X.~Zhu, H.~Hu, S.~Lin, and J.~Dai.
\newblock Deformable convnets v2: More deformable, better results.
\newblock In {\em 2019 IEEE/CVF Conference on Computer Vision and Pattern
  Recognition (CVPR)}, pages 9300--9308, Los Alamitos, CA, USA, Jun 2019. IEEE
  Computer Society.

\end{thebibliography}
\balance

\end{document}